%% file: main.tex
\documentclass[letterpaper]{article} 
\usepackage[preprint]{aaai2027}
\usepackage[hyphens]{url}  
\usepackage{graphicx} 
\usepackage{natbib}  
\usepackage{caption} 
\usepackage{amsmath}
\usepackage{amssymb}
\usepackage{booktabs}
\usepackage{tabularx}
\usepackage{array}
\usepackage{xcolor}
\usepackage{listings}
\usepackage{tcolorbox}
\usepackage{etoolbox}
\tcbuselibrary{listings}

\newcolumntype{Y}{>{\raggedright\arraybackslash}X}
\newcommand{\NA}{--}

\newcommand{\slot}[2]{%
  \hspace{\fontdimen2\font}%
  \(\overset{\text{\scriptsize\sffamily #2}}{\mathtt{\{#1\}}}\)%
}
\newcommand{\slotword}[2]{%
  \slot{#1}{#2}\hspace{\fontdimen2\font}%
}

\lstdefinestyle{constraint}{
  breakindent=0pt,
  breakautoindent=false,
  breakatwhitespace=true
}

\lstdefinestyle{constrainttemplate}{
  style=constraint,
  escapeinside={(*@}{@*)}
}

\newtcblisting{constrainttemplatebox}{
  listing only,
  listing engine=listings,
  colback=black!3,
  colframe=black!30,
  boxrule=0.4pt,
  sharp corners,
  boxsep=0pt,
  left=0.5em,
  right=0.5em,
  top=0.5pt,
  bottom=0.5pt,
  before skip=0pt,
  after skip=0pt,
  before={\par\addvspace{3pt}\nopagebreak[4]\noindent},
  after={\par\addvspace{3pt}},
  listing options={
    style=constrainttemplate,
    frame=none,
    backgroundcolor=\color{black!3},
    xleftmargin=0pt,
    xrightmargin=0pt
  }
}

\newenvironment{constrainttemplatepair}{%
  \par\noindent\begin{minipage}{\linewidth}%
}{%
  \end{minipage}\par
}

\BeforeBeginEnvironment{lstlisting}{%
  \par\nopagebreak[4]\noindent\begin{minipage}{\linewidth}%
}
\AfterEndEnvironment{lstlisting}{%
  \end{minipage}\par
}

\title{LangNCO: Compiling Natural-Language Constraints into Inference-Time Interventions for Neural Combinatorial Optimization}
\author{
    Hironori Ohigashi
}
\affiliations{
    Panasonic Connect Co., Ltd.\\
    ohigashi.hironori@jp.panasonic.com
}

\begin{document}

\maketitle

\begin{abstract}
Neural combinatorial optimization (NCO) rapidly produces
high-quality solutions for problem structures encountered during
training, but operational requirements may change after deployment
and be communicated in natural language. We propose LangNCO, a
framework that compiles instance-specific requirements into
structured inference-time interventions for a frozen NCO solver,
without changing the pretrained policy's architecture, learned input
representation, or parameters. A pretrained LLM serves as a replaceable semantic front
end, translating each requirement into a schema-validated
domain-specific language (DSL) rule set. A deterministic Rule
Executor applies the rules as nonexpansive hard action masks.
Because local masking prevents immediate violations but does not
ensure that a complete feasible continuation remains reachable,
Critical-Offset Decoding extracts rule-referenced critical nodes
and reallocates part of a fixed rollout budget to multi-strength
biased rollouts. Using controlled task-specific language to isolate
solver adaptation from open-ended language robustness, we evaluate
a CVRP policy on road closures, time windows, vehicle assignments,
visit-order requirements, all pairwise combinations, and twelve
additional constraint types. LangNCO achieves 100\% feasibility on
the four primary datasets with mean reference gaps of
1.63--3.26\%, 97.33--100\% feasibility when separately generated
rules are composed for pairwise constraints, and 98.67\% feasibility
across the additional constraints. These
results demonstrate modular post-training adaptation of a frozen
NCO solver through inspectable language-derived rules, without
retraining.
\end{abstract}

\section{Introduction}

Combinatorial optimization underlies routing, scheduling,
production planning, and robotic decision making. In these
domains, high-quality discrete decisions often must be produced
repeatedly under tight time limits. 
Neural combinatorial optimization (NCO) learns models that accelerate solution generation for such problems on new instances
\citep{bello2017neural,bengio2021machine}.
A central line of NCO research focuses on constructive policies
that directly build complete solutions under problem structures
specified during training, enabling fast inference after training
\citep{vinyals2015pointer,nazari2018reinforcement,
kool2019attention,kwon2020pomo}.
In routing, these policies typically construct routes autoregressively by repeatedly selecting the next customer.

Operational requirements, however, often change after deployment. Roads may close, customer time requirements may change, or temporary vehicle-assignment and visit-order rules may be introduced. Natural language provides a flexible interface for specifying such instance-specific requirements, but a standard pretrained solver cannot interpret constraints absent from its training-time problem definition. Retraining the policy, adding constraint-specific inputs, or redesigning the decoder for every transient requirement is impractical.

We study inference-time adaptation to \emph{state-action-level constraints}, whose effects can be evaluated during decoding from the current partial solution, candidate attributes, and exposed environment state. 
This class covers restrictions involving transitions, time,
vehicle identity, visit history, capacity, depot access, and
other state-dependent conditions.
Constraints requiring evaluation only after all routes are complete are outside this scope. Starting from a policy trained only on the standard capacitated vehicle routing problem (CVRP), our goal is to incorporate instance-specific natural-language requirements without constraint-specific policy inputs, architecture changes, or weight updates.

Recent multi-task and unified NCO systems generalize across routing variants through shared architectures and attribute representations
\citep{liu2024multitask,zhou2024mvmoe,berto2025routefinder,li2025cada,goh2025shield,zhou2026urs}.
Their zero-shot settings nevertheless assume that the relevant attributes, state interfaces, and feasibility mechanisms are defined before training; the unseen element is typically a problem variant or a combination of supported attributes within that interface. We instead address post-training requirements first specified in natural language for a concrete instance. LLM-based optimization methods have also generated solutions, mathematical formulations, solver programs, and heuristics
\citep{yang2024opro,liu2024lmea,ahmaditeshnizi2024optimus,jiang2025droc,jiang2025llmcosolver,romeraparedes2024funsearch}.
Rather than assigning combinatorial search to the LLM or reconstructing the base problem through a separate solver pipeline, we retain the pretrained NCO policy as the search engine and use the LLM only to specify state-dependent action interventions.

To realize this division of labor, we propose \emph{LangNCO},
an inference-time framework in which an LLM, serving as a
replaceable semantic front end, translates each requirement
into a compact, schema-validatable domain-specific language
(DSL) that specifies state-dependent changes to the feasible
action set. The validated rules are deterministically compiled, and a Rule
Executor applies them as action masks over the frozen decoder's
base feasibility mask. The NCO policy remains
responsible for route construction and cost minimization. This
design preserves the routing policy's inductive bias and fast
inference, makes interventions inspectable, and avoids free-form
tensor code generation.

Hard masking ensures immediate action legality but not completion viability: a legal choice may consume capacity needed by an assigned customer, delay a time-window customer beyond recovery, or close a route before a prerequisite is served. We therefore introduce \emph{Critical-Offset Decoding}, which extracts explicitly referenced critical nodes from the generated rules and reallocates part of a fixed rollout budget to multi-strength biased rollouts. It never revives a masked action and balances completion-oriented exploration with the original policy's cost-oriented search.

Our contributions are threefold. First, we formulate
post-training request adaptation of a fixed CVRP policy and
introduce a modular language-to-rule interface whose validated
output drives deterministic, nonexpansive action intervention,
without adding constraint-specific inputs to the NCO policy or
changing its architecture or weights. Second, we propose
Critical-Offset Decoding, which biases a subset of rollouts toward
rule-derived critical nodes, reducing future deadlocks and improving
solution quality without increasing the rollout budget. 
Third, we evaluate the framework on four primary constraint
types, all pairwise combinations, and twelve additional types. 
The complete method attains 100\% feasibility on all four
primary datasets with mean reference gaps of 1.63--3.26\%,
97.33--100\% feasibility when separately generated rules are
composed for pairwise constraints, and 98.67\% feasibility across
the additional constraints. These results show
that structured language-derived rules can extend a frozen NCO
solver beyond its training-time problem definition without
retraining.

\section{Related Work}

\paragraph{NCO and generalization.}
Constructive NCO policies build solutions sequentially. Pointer Networks introduced attention-based permutation generation, while subsequent reinforcement-learning methods learned routing policies without optimal labels; the Attention Model and POMO further established Transformer-style encoders, feasibility masking, and multistart decoding \citep{vinyals2015pointer,bello2017neural,nazari2018reinforcement,kool2019attention,kwon2020pomo}. Later work improved scalability and decoder capacity and systematized NCO evaluation \citep{jin2023pointerformer,luo2023heavy,bengio2021machine,berto2025rl4co}. Generalist solvers instead share representations across routing problems, constraint combinations, and distributions \citep{liu2024multitask,zhou2024mvmoe,berto2025routefinder,li2025cada,goh2025shield}; URS extends this direction with unified representations and model-agnostic feasibility masks for unseen routing variants \citep{zhou2026urs}, while other work studies distributional and size generalization \citep{jiang2022distributionally,zhou2023omni}. These methods generalize within structured solver interfaces whose attributes and feasibility mechanisms are specified before training. In contrast, we freeze a policy trained only on standard CVRP and compose concrete target entities, state guards, and action effects from natural language at inference time. Our setting is therefore post-training request adaptation rather than cross-problem generalization of a pretrained generalist.

\paragraph{LLMs for combinatorial optimization.}
LLMs have been used to propose complete or intermediate solutions \citep{yang2024opro,liu2024lmea,iklassov2024selfguiding,jiang2025llmcosolver}, translate natural-language descriptions into optimization formulations or solver programs \citep{ahmaditeshnizi2024optimus,zhang2024optllm,jiang2025droc,astorga2025autoformulation,ramamonjison2022nl4opt}, and search for reusable algorithms or heuristics \citep{romeraparedes2024funsearch,ye2024reevo}. These approaches either assign substantial combinatorial search to the LLM or construct a separate optimization pipeline. We instead generate only a compact action-intervention rule and retain the pretrained NCO policy as the route-construction and search component. The base CVRP formulation remains unchanged, while the request modifies only the admissibility or priority of affected actions. This separation keeps semantic interpretation distinct from combinatorial search for a low-cost feasible solution.

\paragraph{Constraint-aware decoding and structured intervention.}
Action masks are standard in autoregressive routing, and related state-specific masks and safety layers suppress invalid or unsafe actions \citep{wang2024masks,krasowski2023safe}. Proactive Infeasibility Prevention masks actions using constraint-specific lookahead, while LazyMask combines lookahead with backtracking \citep{bi2024pip,li2026lmask}. Our intervention is instead derived from an instance-specific language-generated rule and applied to a frozen standard-CVRP policy. The same representation provides both a deterministic hard mask for local legality and a structural signal that directs inference-time search toward critical nodes.

Constrained generation methods such as PICARD, grammar-constrained decoding, and grammar prompting improve formal validity \citep{scholak2021picard,geng2023grammar,wang2023grammar}, whereas code-generation studies show that syntactic validity does not prevent failures in API use, data shapes, or executable semantics \citep{chen2021codex,li2022alphacode,lai2023ds1000,ni2024l2ceval}. Our DSL therefore restricts generation to guarded target/effect rules, while parsing, schema validation, compilation, and validation-feedback retries fix tensor operations, shape handling, and mask-composition semantics outside the LLM. The resulting rules remain inspectable, and their structured fields are reused for decoding bias. Our contribution is thus a shared intermediate representation connecting natural-language interpretation, deterministic action intervention, and inference-time search guidance for a fixed learned optimizer.

\section{Method}

\subsection{Problem Setting}

Let $x$ be a capacitated vehicle routing problem (CVRP) instance on nodes
$V=\{0,1,\ldots,n\}$, where node $0$ is the depot, customer $i$ has demand
$d_i$, and vehicle capacity is $Q$. A solution $\tau$ consists of
depot-to-depot routes that visit every customer exactly once without
exceeding capacity. With edge cost $c_{ij}$, its travel cost is
$C(\tau\mid x)$.

A pretrained autoregressive NCO policy $\pi_\theta$ selects the next node
$a_t$ from a partial state $s_t$. Its parameters $\theta$ remain fixed
throughout this work.

At inference time, the target instance is accompanied by a natural-language
specification $\ell$, which may contain one or more requirements. Let
$\Omega(x,\ell)$ denote the set of solutions satisfying the base CVRP
constraints and the requirements expressed by $\ell$. The ideal constrained objective is
\begin{equation}
    \tau^\star
    \in
    \arg\min_{\tau\in\Omega(x,\ell)}
    C(\tau\mid x).
    \label{eq:task-objective}
\end{equation}

The LLM rule generator maps $\ell$ to a validated DSL rule set
$r=G(\ell)$. The set $\Omega(x,\ell)$ is a conceptual definition of the
target solutions and is not available to the solver. During inference, the
solver operates only on the generated rule set $r$.

\begin{figure*}[t]
    \centering
    \includegraphics[width=0.98\textwidth]{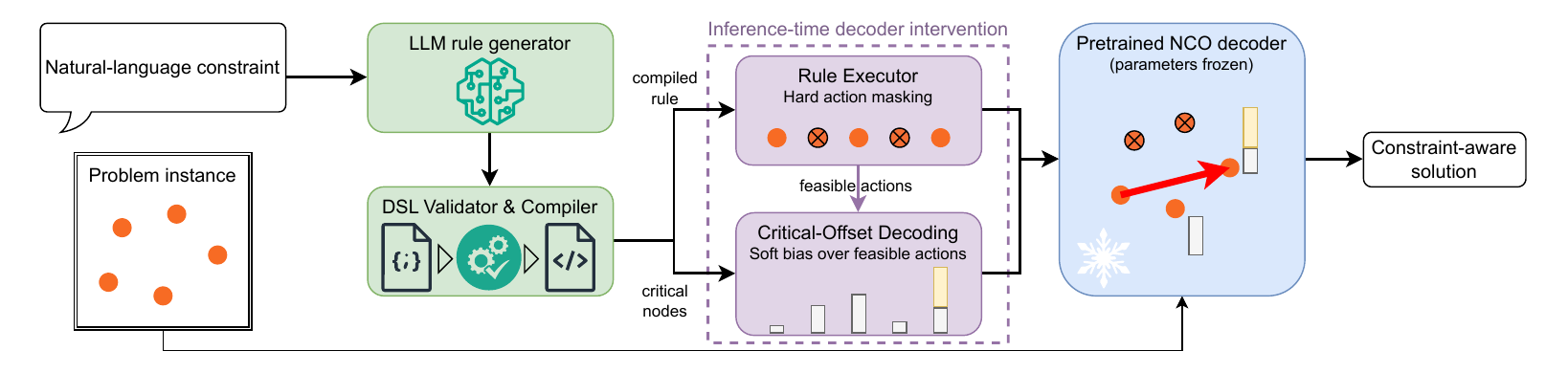}
    \caption{Inference-time intervention. A natural-language specification is
    translated into a DSL rule set that is validated and compiled to drive
    hard action masking and critical-node-biased rollouts. The pretrained NCO
    policy's parameters and learned input representation remain unchanged.}
    \label{fig:overview}
\end{figure*}

The base decoder applies a CVRP feasibility mask
$m_t^{\mathrm{base}}(a)$, where a value of one denotes an admissible action.
We derive an additional rule mask $m_t^{\mathrm{rule}}(a,r)$ and apply
\begin{equation}
    m_t(a,r)
    =
    m_t^{\mathrm{base}}(a)
    \land
    m_t^{\mathrm{rule}}(a,r).
    \label{eq:combined-mask}
\end{equation}
Thus, a generated rule may remove a base-feasible action but cannot
re-enable an action prohibited by the CVRP environment.
Figure~\ref{fig:overview} summarizes the resulting inference-time
intervention pipeline.

\subsection{Compiling Natural-Language Constraints into State--Action Rules}

Rather than generating a final route or free-form code, the LLM
translates each requirement into one or more instance-specific
state--action rules in a JSON DSL. Each rule specifies candidate
targets, optional state- and candidate-level conditions, and one of
two nonexpansive effects: blocking matched actions or retaining only
them.

Targets may denote the depot, all customers, fixed node IDs, or
attribute-defined customer sets. Conditions may reference the current
node, cumulative or vehicle-specific visit history, vehicle identity,
capacity, route length, demand, coordinates, and auxiliary observations
exposed by the simulator.

The DSL fixes typed operators and state-access forms while allowing an
extensible observation namespace. At inference time, the simulator may
expose instance-specific attributes or pairwise matrices, such as elapsed
time or travel times. These observations are used only by the Rule
Executor and are not added to the pretrained NCO policy input.

Rather than selecting from predefined VRP-specific rules or DSL templates,
the system composes targets, constants, guards, and effects from shared
primitives and available observations. A request is supported when its
required state quantities are exposed. Restricting generation to
system-defined operations keeps execution semantics under compiler control.

For a constraint $c$, let $T_c$ be its target set, $g_c(s_t)$ its
state-level condition, and $h_c(s_t,a)$ its candidate-level condition.
An omitted condition evaluates to true. The matched actions are
\begin{equation}
    M_c(s_t)
    =
    \left\{
    a\in T_c
    \,\middle|\,
    g_c(s_t)\land h_c(s_t,a)
    \right\}.
    \label{eq:matched-set}
\end{equation}
Given the current feasible set $\mathcal A$, a constraint applies one of
\begin{equation}
    \mathcal A
    \leftarrow
    \mathcal A\setminus M_c(s_t)
    \qquad\text{or}\qquad
    \mathcal A
    \leftarrow
    \mathcal A\cap M_c(s_t).
    \label{eq:dsl-effects}
\end{equation}
Both operations are nonexpansive, so composing multiple constraints
preserves
\begin{equation}
    \mathcal A_t^{\mathrm{final}}
    \subseteq
    \mathcal A_t^{\mathrm{base}}.
    \label{eq:nonexpansive-dsl}
\end{equation}
A single natural-language requirement may compile into multiple
constraints. For example, an undirected road closure is represented by two
symmetric transition constraints.

\paragraph{Validation and compilation.}
After JSON parsing, the generated rule set is validated against the DSL
schema and checked by compilation using the same compiler that generates
the executable rules. Schema validation checks the rule-set structure and
schema-defined expression forms. The compile check rejects expressions that
cannot be translated under the supported DSL semantics, including
unsupported operators, access patterns, and node-set forms, as well as
candidate-indexed references used outside a candidate context.

For each valid global or instance-scoped collection of constraints, the
compiler emits a self-contained PyTorch mask function. It translates target
sets into batched candidate masks, state-level conditions into row-wise
expressions, and candidate-level conditions into node-wise expressions.
Where needed, the compiler reconciles original-unit quantities in the DSL
with the normalized solver state and fixes state access, broadcasting,
tensor-shape handling, and nonexpansive mask composition.

If JSON parsing, schema validation, or compilation fails, the resulting
error is returned to the LLM for at most two repair attempts. These checks
ensure that a rule is schema-conformant and executable under the supported
DSL, but do not verify semantic agreement with the natural-language
requirement. Additional details on the typed rule representation,
validation procedure, and compilation semantics are provided in the
Technical Supplement.

\subsection{Rule Executor}

At decoding step $t$, the Rule Executor evaluates all compiled constraints
for the target instance against the exposed routing state and candidate
actions. The state interface contains routing quantities rather than neural
hidden representations, including the current node, visit histories,
current vehicle, capacity usage, route length, time, coordinates, demand,
and instance-specific auxiliary observations.

The constraints are therefore independent of the internal architecture that
produces the action logits, provided that the required state interface is
exposed. If the NCO decoder produces logits $z_t(a)$, the combined mask
gives
\begin{equation}
    \widetilde z_t(a)
    =
    \begin{cases}
        z_t(a), & m_t(a,r)=1,\\
        -\infty, & m_t(a,r)=0.
    \end{cases}
    \label{eq:masked-logits}
\end{equation}
The same executor can be used with greedy or sampling-based decoding.

To avoid a degenerate all-false mask, the executor skips a
retain-only update whose target contains neither the depot nor
an unvisited customer; this implementation safeguard leaves the
current mask unchanged but does not guarantee satisfaction of
the skipped constraint.

A rollout is recorded as deadlocked when unserved customers remain but the
executed mask contains no available action. The method neither backtracks
from a failed prefix nor regenerates its rules. For $R$ compiled constraints
and $|V|$ candidate actions, vectorized rule evaluation costs
$O(R|V|)$ per decoding step.

\subsection{Critical-Offset Decoding}
\label{sec:critical-offset}

\paragraph{From local legality to completion-oriented search.}
Hard masking guarantees only local legality with respect to the executed
base and DSL constraints: an allowed action may still eliminate all
complete continuations. For example, delaying a time-window customer may
make its deadline unreachable, another customer may consume the capacity
needed by an assigned customer, or a route may terminate before its ordering
requirements are completed.

Exactly determining whether each action preserves a complete continuation
would require constraint-specific lookahead or solving a constrained
completion problem for every candidate. Critical-Offset Decoding instead
derives a low-cost, completion-oriented search prior from the structure of
the generated DSL.

Many such nonlocal failures are associated with nodes referenced by the
generated constraints. A customer directly restricted by a constraint may
lose its service opportunity if it is postponed until a deadline expires,
the designated vehicle loses capacity, or the current route terminates.
Likewise, the state of a node referenced by a constraint condition may
determine whether another customer becomes or remains available.

\paragraph{Subject and condition nodes.}
We call a node identified from a constraint target a \emph{subject node},
because the constraint directly changes its availability. We call a node
identified from a constraint condition a \emph{condition node}, because its
state can affect the future availability of a subject node.

Let $N_{\mathrm{tar}}(c)$ and $N_{\mathrm{cond}}(c)$ denote the node IDs
resolved from the target and conditions of constraint $c$, respectively.
For a rule set $r$, we extract
\begin{align}
    \mathcal S_r
    &=
    \bigcup_{c\in r} N_{\mathrm{tar}}(c),
    \nonumber\\
    \mathcal G_r
    &=
    \bigcup_{c\in r} N_{\mathrm{cond}}(c),
    \nonumber\\
    \mathcal C_r
    &=
    (\mathcal S_r\cup\mathcal G_r)\setminus\{0\},
    \label{eq:critical-set}
\end{align}
where $\mathcal S_r$, $\mathcal G_r$, and $\mathcal C_r$ are the subject,
condition, and critical-node sets, respectively. We exclude the depot in all
experiments because prioritizing it may encourage premature route
termination.

The current implementation uses a lightweight syntax-directed extractor for
statically resolvable node references. Supporting additional typed node
accessors or resolving attribute-defined targets would require only
extending this extraction step; the Rule Executor and Critical-Offset
Decoding would remain unchanged.

Critical-node extraction converts the compositional structure of the
generated DSL into a shared, completion-oriented search prior. Subject nodes
identify actions whose future service opportunities may be lost, while
condition nodes identify state changes that affect their future
availability. Extraction is deterministic and requires neither a
constraint-type label, a dataset annotation, nor an additional LLM call.

\paragraph{State-dependent critical actions.}
The extracted set $\mathcal C_r$ remains fixed throughout a rollout, but
only currently legal critical nodes receive an offset. After applying the
base and rule masks, let
\begin{equation}
    \mathcal A_t
    =
    \{a\in V\mid m_t(a,r)=1\}
    \label{eq:admissible-actions}
\end{equation}
be the locally admissible actions, and define
\begin{equation}
    \mathcal C_t
    =
    \mathcal C_r\cap\mathcal A_t.
    \label{eq:active-critical-set}
\end{equation}
Thus, the active critical set changes with the current state and hard mask,
even though its DSL-derived source set is static.

Consider a same-route ordering constraint requiring every node in a
prerequisite set $B$ to be served before a subject node $j$. While the
prerequisites remain incomplete, the Rule Executor masks $j$, so the offset
cannot make it illegally selectable and instead acts on feasible condition
nodes in $B$. After the prerequisites are completed, $j$ is no longer
blocked by the ordering constraint and may become feasible and receive
priority. Similarly, a time-window or vehicle-assigned subject receives
additional weight only while it remains legal.

\paragraph{Offset-guided action selection.}
Let
\begin{equation}
    p_t(a)
    =
    \operatorname{softmax}(\widetilde z_t)_a
    \label{eq:masked-policy-probability}
\end{equation}
denote the policy probability after hard masking, with $p_t(a)=0$ for
$a\notin\mathcal A_t$. For offset strength $\alpha\in[0,1]$,
Critical-Offset Decoding defines the nonnormalized action weight
\begin{equation}
    q_t^{(\alpha)}(a)
    =
    \begin{cases}
        (1-\alpha)p_t(a)+\alpha,
        & a\in\mathcal C_t,\\
        p_t(a),
        & \text{otherwise}.
    \end{cases}
    \label{eq:critical-offset}
\end{equation}
When $\alpha=0$, the original masked policy is recovered exactly.
Increasing $\alpha$ assigns greater aggregate selection weight to currently
feasible critical nodes and reduces policy-induced differences among them.

For greedy decoding, the selected action is
\begin{equation}
    a_t
    =
    \arg\max_{a\in\mathcal A_t}
    q_t^{(\alpha)}(a).
    \label{eq:offset-greedy}
\end{equation}
For sampling, the weights are normalized over the admissible actions:
\begin{equation}
    \bar q_t^{(\alpha)}(a)
    =
    \frac{q_t^{(\alpha)}(a)}
    {\sum_{a'\in\mathcal A_t}q_t^{(\alpha)}(a')}.
    \label{eq:offset-sampling}
\end{equation}
At $\alpha=1$, each feasible critical node receives an unnormalized weight
of one. Under sampling, this strongly prioritizes critical nodes but does
not strictly force their selection because noncritical actions retain their
original positive weights.

Crucially, the offset is applied only to
$\mathcal C_t=\mathcal C_r\cap\mathcal A_t$. It therefore cannot assign
positive weight to an action prohibited by either the base or rule mask.
The hard mask determines legality, whereas Critical-Offset Decoding changes
only the preference among actions that remain legal.

\paragraph{Multiple offset strengths.}
The useful offset strength depends on the instance geometry, capacities,
deadlines, route state, and original policy distribution. Strong priority
can preserve a scarce service opportunity, but may also cause an unnecessary
detour or serve a rule-related node earlier than needed. We therefore treat
$\alpha$ as a rollout dimension rather than selecting one global value.

For each POMO start and geometric augmentation, the total sampling budget
remains 100 rollouts. Ninety use ordinary sampling with $\alpha=0$, and the
remaining ten are divided equally across
\begin{equation}
    \Lambda
    =
    \{0.2,0.4,0.6,0.8,1.0\},
    \label{eq:offset-strengths}
\end{equation}
giving two rollouts per nonzero strength. Equivalently,
\begin{equation}
    B_0=90,
    \qquad
    B_\alpha=2\ \ (\alpha\in\Lambda),
    \qquad
    B_0+\sum_{\alpha\in\Lambda}B_\alpha=100.
    \label{eq:rollout-allocation}
\end{equation}
All rollouts use the same frozen NCO network and generated rule set.
Ordinary rollouts preserve the original policy's cost-oriented search,
while offset rollouts explore several strengths of priority toward
rule-related nodes.

\paragraph{Rollout selection and scope.}
Let $\tau_b$ be the route produced by rollout $b$, and let
$\mathcal B_{\mathrm{comp}}$ denote the rollouts that reach a complete
base-CVRP terminal state. Incomplete and deadlocked rollouts are discarded,
and completed rollouts are ranked solely by travel cost:
\begin{equation}
    b^\star
    \in
    \arg\min_{b\in\mathcal{B}_{\mathrm{comp}}}
    C(\tau_b\mid x),
    \qquad
    \tau^\star = \tau_{b^\star}.
    \label{eq:rollout-selection}
\end{equation}
If $\mathcal B_{\mathrm{comp}}=\emptyset$, the solve is recorded as
unsuccessful; this is not an infeasibility certificate for the constrained
instance.

Critical-Offset Decoding is a rule-derived rollout prior rather than an
exact viability certificate. It neither simulates every successor action
nor backtracks from a failed prefix. Its effect also depends on the
constraint structure: locally enforced transition restrictions such as road
closures may require little additional directional guidance. Retaining
mostly ordinary rollouts and using multiple offset strengths limits the
effect of unnecessary directional priority.

Because Equation~\ref{eq:critical-offset} modifies only actions satisfying
$m_t(a,r)=1$, the hard feasible set remains unchanged. Critical weighting
adds $O(|V|)$ work per decoding step; together with vectorized evaluation of
$R$ compiled constraints, the intervention costs $O(R|V|)$ per step. The
number of policy evaluations remains identical to the baseline with 100
rollouts per POMO start and augmentation.

\section{Experiments}

\subsection{Experimental Setup}

\paragraph{Base solver and data.}
We use RouteFinder \citep{berto2025routefinder} trained only on standard CVRP. The base test set contains 300 instances from the LLMCoSolver benchmark \citep{jiang2025llmcosolver}, with 10--100 customer nodes and LKH-3 near-optimal routes supplied for the unconstrained instances. Decoding uses POMO-style multistarts and eight geometric augmentations. The RouteFinder architecture and parameters remain fixed in all experiments.

\paragraph{Constraint-augmented datasets.}
We construct four primary datasets by adding one hard constraint to each base instance and selecting its targets so that the original high-quality solution violates the new requirement. \textsc{Road} closes an adjacent non-depot edge in both directions. \textsc{Time} imposes a no-wait hard time window of width 18--20 on an early-route customer, using travel time $\lceil 0.1 \times \text{Euclidean distance} \rceil$. \textsc{Vehicle} assigns two capacity-compatible, high-demand customers, selected from different reference routes when possible, to the same vehicle. \textsc{Order} requires two prerequisite nodes to precede a target node on the same route, using a sandwich construction such that neither the original route nor its reversal satisfies the precedence constraint. Every retained primary instance has a constraint-feasible reference solution.

The constraints are expressed using controlled, task-specific
natural-language descriptions with instance-specific entities and
values. Our evaluation focuses on constraint-type and compositional
breadth rather than systematic robustness to paraphrasing.

\paragraph{Methods and metrics.}
\textsc{LLMCoSolver} and \textsc{Direct GPT-5} serve as direct-solution baselines that generate complete routes from the problem description. \textsc{NCO w/o Intervention} is the unmodified pretrained RouteFinder decoder; it receives neither the natural-language constraint nor a derived rule mask. Our variants use either Gemma4 (\texttt{gemma4:31b-it-q8\_0}) or GPT-5 only to generate the DSL, after which both use the same Rule Executor and Critical-Offset Decoding.

DSL generation uses an initial attempt and up to two validation-feedback retries after parsing, schema-validation, or compile-check failures. If no valid rule is obtained, the solver proceeds without an additional rule. After inference, an independently implemented evaluation-only constraint checker determines whether each returned route satisfies the base CVRP constraints and all applicable requirements in the natural-language specification. All feasibility rates reported throughout the experiments are computed using this post-hoc checker. The checker is not exposed to the inference pipeline and cannot affect rule generation, validation-feedback retries, action masking, decoding, completion filtering, or cost-based rollout selection. Unparseable route outputs, runtime failures, and decoding deadlocks count as infeasible. For feasible outputs, the reference gap is $(C-C_{\mathrm{ref}})/C_{\mathrm{ref}}\times100$. 
Primary-task references use LKH-3 for Road \citep{helsgaun2017lkh3}
and OR-Tools for Time, Vehicle, and Order \citep{ortools}; pairwise
references also use OR-Tools. None provides an optimality certificate.

The Technical Supplement provides additional implementation details,
model versions and inference settings, solver configurations and time
limits, random seeds, hardware details, and complete experimental results.

\subsection{End-to-End Results}

\begin{table*}[t]
\caption{
End-to-end performance.
Gap is computed over feasible outputs only; 
``--'' denotes no feasible output.
}
\label{tab:e2e}
\centering
\small
\setlength{\tabcolsep}{3.8pt}
\begin{tabular}{l *{4}{rr}}
\toprule
& \multicolumn{2}{c}{Road}
& \multicolumn{2}{c}{Time}
& \multicolumn{2}{c}{Vehicle}
& \multicolumn{2}{c}{Order} \\
\cmidrule(lr){2-3}
\cmidrule(lr){4-5}
\cmidrule(lr){6-7}
\cmidrule(lr){8-9}
Method
& Feas.$\uparrow$ & Gap$\downarrow$
& Feas.$\uparrow$ & Gap$\downarrow$
& Feas.$\uparrow$ & Gap$\downarrow$
& Feas.$\uparrow$ & Gap$\downarrow$ \\
\midrule

LLMCoSolver
& 36.33\% & 5.70\%
& 13.67\% & 4.42\%
& 2.00\%  & 0.58\%
& 0.00\%  & -- \\

Direct GPT-5
& 95.00\% & 35.02\%
& 79.67\% & 120.01\%
& 95.33\% & 40.39\%
& 95.67\% & 32.04\% \\

NCO w/o intervention
& 37.33\% & 2.65\%
& 11.00\% & 1.63\%
& 1.00\%  & 2.56\%
& 1.33\%  & 1.47\% \\

\midrule

\textbf{LangNCO-Gemma4}
& \textbf{100.00\%} & \textbf{1.63\%}
& \textbf{100.00\%} & \textbf{1.89\%}
& \textbf{100.00\%} & \textbf{2.53\%}
& \textbf{100.00\%} & 3.26\% \\

\textbf{LangNCO-GPT-5}
& \textbf{100.00\%} & \textbf{1.63\%}
& \textbf{100.00\%} & \textbf{1.89\%}
& \textbf{100.00\%} & 2.68\%
& \textbf{100.00\%} & \textbf{2.97\%} \\

\bottomrule
\end{tabular}
\end{table*}

Table~\ref{tab:e2e} evaluates the complete language-to-route pipeline. Both proposed variants achieve 100\% feasibility on all four datasets, with reference gaps of 1.63--3.26\%. In contrast, NCO without intervention retains low route costs only on the few instances that accidentally satisfy the new constraint, reaching just 1.00--37.33\% feasibility. Direct GPT-5 more often satisfies the constraints but produces substantially worse routes, with gaps of 32.04--120.01\%. LLMCoSolver reaches at most 13.67\% feasibility on three of the four unseen constraint types. These results support assigning language interpretation to the LLM while retaining the pretrained NCO policy as the combinatorial search engine.

\paragraph{Rule-representation robustness.}
As an auxiliary comparison, we replace the DSL with free-form Python rule generation while keeping the downstream solver and decoding strategy fixed. The DSL achieves 100\% feasibility in all eight generator--constraint settings. Python generation falls below 100\% in three settings, with a minimum feasibility of 89.33\%. Observed failures include tensor-shape mismatches, inconsistent device placement, and indexing operations that modify a copy rather than the original action mask. This auxiliary comparison indicates that restricting rule generation to the DSL can mitigate low-level implementation failures observed with free-form Python. A per-task breakdown and representative generated-code failures are included in the Technical Supplement.

\subsection{Effectiveness of Critical-Offset Decoding}
\label{sec:offset-results}

To isolate rollout guidance from errors in generated hard masks,
we replace the generated masks with independently implemented
reference masks while retaining DSL-derived critical-node extraction. 
Critical nodes are still extracted from the DSL generated by LangNCO-Gemma4 and are not supplied through dataset labels. 
The greedy comparison holds the original greedy decoding fixed and compares
one additional ordinary sample with one additional offset-guided rollout.
For sampling, both standard sampling and Critical-Offset Decoding use the
same budget of 100 rollouts per POMO start and augmentation.

\begin{table*}[t]
\caption{
Effect of Critical-Offset Decoding.
Gap is computed over each strategy's feasible outputs only.
}
\label{tab:offset}
\centering
\small
\setlength{\tabcolsep}{3.5pt}
\begin{tabular}{l *{4}{rr}}
\toprule
& \multicolumn{2}{c}{Road}
& \multicolumn{2}{c}{Time}
& \multicolumn{2}{c}{Vehicle}
& \multicolumn{2}{c}{Order} \\
\cmidrule(lr){2-3}
\cmidrule(lr){4-5}
\cmidrule(lr){6-7}
\cmidrule(lr){8-9}
Strategy
& Feas.$\uparrow$ & Gap$\downarrow$
& Feas.$\uparrow$ & Gap$\downarrow$
& Feas.$\uparrow$ & Gap$\downarrow$
& Feas.$\uparrow$ & Gap$\downarrow$ \\
\midrule

NCO greedy
& 100.00\% & 3.19\%
& 99.67\%  & 4.47\%
& 75.33\%  & 5.64\%
& 75.00\%  & 7.69\% \\

NCO greedy + 1 sample
& 100.00\% & 2.28\%
& 99.67\%  & 3.83\%
& 77.33\%  & 5.26\%
& 83.67\%  & 6.56\% \\

\textbf{Critical-Offset greedy}
& 100.00\% & 3.15\%
& 100.00\% & 3.52\%
& 100.00\% & 4.56\%
& 100.00\% & 6.81\% \\

\midrule

NCO sampling
& 100.00\% & 1.64\%
& 99.67\%  & 2.19\%
& 87.67\%  & 3.33\%
& 97.00\%  & 3.52\% \\

\textbf{Critical-Offset sampling}
& \textbf{100.00\%} & \textbf{1.63\%}
& \textbf{100.00\%} & \textbf{1.89\%}
& \textbf{100.00\%} & \textbf{2.53\%}
& \textbf{100.00\%} & \textbf{2.85\%} \\

\bottomrule
\end{tabular}
\end{table*}

The greedy control shows that the gain cannot be explained by adding one more rollout alone. On \textsc{Vehicle}, adding one ordinary sample raises feasibility only from 75.33\% to 77.33\%, whereas replacing that additional rollout with an offset-guided rollout raises it to 100\%. On \textsc{Order}, the corresponding feasibility values are 83.67\% and 100\%.

Under the same budget of 100 rollouts per POMO start and augmentation, standard sampling fails to produce feasible solutions on 0.33\%, 12.33\%, and 3.00\% of the \textsc{Time}, \textsc{Vehicle}, and \textsc{Order} instances, respectively, whereas Critical-Offset Decoding reaches 100\% feasibility on all four datasets. The reported mean gaps are also lower, changing from 2.19\%, 3.33\%, and 3.52\% to 1.89\%, 2.53\%, and 2.85\%, respectively. For this matched-budget comparison, the mean decoder time is nearly unchanged, increasing from 173.78 to 174.68 ms; language generation and rule compilation are excluded from this measurement.

These results reflect the distinction between local legality and completion viability described in Section~\ref{sec:critical-offset}. Ordinary sampling may postpone a currently feasible but low-probability critical node until a deadline expires, the designated vehicle's capacity is consumed, or the current route is closed. Offset-guided rollouts add trajectories that serve such nodes while the opportunity remains available. 
On \textsc{Road}, the negligible change is expected because the hard mask already captures the immediate violation.

\subsection{Constraint Composition and Breadth}

\paragraph{Pairwise constraint composition.}
We evaluate all six pairwise combinations of the four primary
constraints. The two natural-language constraints are translated
separately, aggregated into a single rule set, and executed without
pair-specific prompts or handlers. As a time-limited reference, we
use OR-Tools with a single initial-solution strategy selected for the
highest aggregate reference coverage among several standard
alternatives. The exact strategy, solver configuration, and time limit
are provided in the Technical Supplement. Table~\ref{tab:composition}
reports the numbers of feasible solutions returned by LangNCO-Gemma4
and the reference solver out of 300 instances. The reference gap is
computed only over instances for which both return feasible solutions.

\begin{table}[t]
\caption{
Pairwise constraint composition for LangNCO-Gemma4.
Counts are Ours / Ref.\ out of 300; gap is computed over jointly feasible instances.
}
\label{tab:composition}
\centering
\small
\begin{tabular*}{\columnwidth}{
  @{\extracolsep{\fill}}lrr@{}
}
\toprule
Combination
& Feasible (Ours / Ref.)
& Gap$\downarrow$ \\
\midrule
Road + Time
& 300 / 300 & 2.25\% \\
Road + Vehicle
& 300 / 300 & 3.14\% \\
Road + Order
& 300 / 300 & 3.94\% \\
Time + Vehicle
& 300 / 300 & 4.01\% \\
Time + Order
& 292 / 289 & 5.18\% \\
Vehicle + Order
& 294 / 297 & 6.77\% \\
\bottomrule
\end{tabular*}
\end{table}

LangNCO-Gemma4 returns feasible solutions for 292--300 of the
300 instances across all six combinations. It covers every
reference-feasible instance in five combinations and 294 of 297
instances for \textsc{Vehicle} + \textsc{Order}. For
\textsc{Time} + \textsc{Order}, it additionally finds feasible
solutions for three instances not solved by the time-limited reference.
Because failure of the reference solver is not an infeasibility
certificate, this indicates complementary solution coverage rather
than superiority over OR-Tools. The reference gaps range from
2.25\% to 6.77\%, with the largest gap occurring for
\textsc{Vehicle} + \textsc{Order}, which jointly restricts route
membership and within-route sequence.

\paragraph{Additional constraint types.}
We further evaluate twelve additional constraint types with 50 instances each: first delivery, visit position, arrival capacity reserve, heavy-delivery limit, minimum service interval, depot closure, small-delivery assignment, return deadline, vehicle workload limit, consecutive deliveries, customer incompatibility, and restricted-area transition. Because these breadth-evaluation datasets do not have constraint-specific reference solutions, we report feasibility only. Without changing the pipeline, LangNCO-Gemma4 obtains 592 feasible solutions out of 600 (98.67\%), compared with 94 out of 600 (15.67\%) for NCO without intervention. Ten constraint types reach 100\% feasibility; small-delivery assignment reaches 96\%, and return deadline reaches 88\%.

These results indicate that the pipeline is not limited to the four primary constraint constructions. Nevertheless, the two lower-feasibility tasks reveal distinct limitations. For small-delivery assignment, the generated DSL is correct, but the critical-node extractor does not expand an attribute-defined target into a fixed node set. For return deadline, the LLM sometimes produces a structurally valid but semantically incorrect rule that restricts customer arrivals without constraining depot return. These cases distinguish search-guidance limitations from semantic rule-generation errors.

\section{Conclusion and Future Work}

We introduced LangNCO, a framework for adapting a frozen NCO
solver to instance-specific natural-language constraints at inference
time. An LLM compiles each requirement into a schema-validated
DSL rule, which a deterministic Rule Executor applies as an action
mask. Critical-Offset Decoding further uses rule-derived critical
nodes to mitigate future deadlocks within a fixed rollout budget.
In the controlled decoding evaluation with independently
implemented reference rule masks, it achieved 100\% feasibility
on all four primary constraints under the same budget of 100 rollouts
per POMO start and augmentation,
compared with 87.67--100\% for standard sampling, while yielding
lower mean reference gaps on the three nonlocal constraints. The
complete LangNCO pipeline achieved 100\% feasibility with mean
reference gaps of 1.63--3.26\% on the primary constraints,
97.33--100\% feasibility when separately generated rules are
composed for all pairwise combinations, and 98.67\% feasibility
across twelve additional constraint types.
These results show that a shared structured representation can
connect natural-language interpretation, deterministic action
intervention, and inference-time search guidance for a pretrained
solver without retraining. Future work will study robustness to
paraphrased and noisier language, semantic rule verification,
attribute-aware critical-node extraction, stronger
completion-aware decoding, and broader optimization domains.

\paragraph{Generative AI Use Disclosure.}
Generative AI tools assisted with editing, translation, restructuring,
implementation discussions, and code completion/generation. The author
reviewed all outputs, tested code where applicable, and takes full
responsibility for the work.

\clearpage

\bibliography{aaai2027}

\clearpage
\appendix
\section*{Technical Supplement}
\input{technical_supplement_body}

\end{document}

%% file: technical_supplement_body.tex
This supplement provides the dataset construction and evaluation protocol,
implementation details, reproducibility settings, complete numerical results,
and failure analyses for the main paper. The post-hoc checker, reference solvers, and all
reported metrics are defined here independently of the LangNCO inference
pipeline. Sections~A--C document dataset construction, the LangNCO
implementation, and reproducibility settings. Section~D reports complete
experimental results, including decoding ablations and a Critical-Offset
sensitivity analysis; Section~E analyzes representative failures and decoding
behavior.

\setcounter{secnumdepth}{3}
\renewcommand{\thesubparagraph}{\thesubsection.\arabic{subparagraph}}

\section{Dataset Construction and Evaluation Protocol}
\label{sec:data}

\subsection{Base and Primary Constraint Datasets}

\subsubsection{Base CVRP benchmark}

We use the public 300-instance capacitated vehicle routing problem (CVRP)
test split released with LLMCoSolver~\citep{jiang2025llmcosolver}. Each
instance contains one depot and 10--100 customers. Using the same numerical
instances permits instance-level comparison between LLMCoSolver, which
generates complete routes from text, and LangNCO, which intervenes in a
standard CVRP policy. The split also spans multiple sizes and spatial
distributions and includes a high-quality LKH-3 route for every instance.

Throughout this supplement, $V=\{0,1,\ldots,n\}$ is the node set and the
universe of decoder actions, with $N=|V|=n+1$ total nodes. Node 0 is the depot
and nodes $1,\ldots,n$ are the $n$ customers. Customer $i$ has raw coordinate
$x_i$, raw demand $d_i$, and the homogeneous vehicle capacity is $Q$. We write
$c_{ij}=\lVert x_i-x_j\rVert_2$ for the unrounded Euclidean edge cost between
nodes $i$ and $j$.

In the released generation procedure, the number of customers is sampled from 
the integers in $[10,100]$. Coordinates are drawn either uniformly or from 
Gaussian mixture distributions parameterized by the number of clusters $c$ 
and scale $l$, using $(c,l)=(2,5)$ and $(3,10)$. Customer demands are sampled 
uniformly from ${1,\ldots,10}$, while the depot is assigned a demand value of 
zero as a representational convention inherited from the original dataset.
In the released implementation, if $\bar d$ is the mean customer demand and 
$r\sim\operatorname{Triangular}(3,6,25)$, the homogeneous vehicle capacity is 
set to $Q=\lfloor r\bar d\rfloor$.
Capacity therefore depends on each instance's demand profile and sampled
coefficient, rather than being fixed by problem size.

The released text-attributed instances contain the fields in
Table~\ref{tab:tai-fields}.

We use these 300 numerical instances and their supplied routes without regeneration 
or resampling. Before supplying the instances to RouteFinder, we divide all coordinate 
values by 1000 to match the normalized coordinate scale expected by the RouteFinder 
implementation. This rescaling is applied only to the RouteFinder input; solution costs 
and reference gaps are evaluated using the original, unnormalized coordinates.
Following the source benchmark, the routes are
treated as near-optimal references rather than optimality-certified ground
truth. Each primary dataset adds one natural-language constraint to each base
instance. The nearest-neighbor features are not part of the RouteFinder state
given to LangNCO or NCO w/o intervention. They are reconstructed from the
coordinates only for the LLMCoSolver baseline to preserve its original input
format.

\subsubsection{Common construction protocol}

For each task, a single fixed sentence template is instantiated with
instance-specific nodes or parameters. Targets are selected from the supplied
reference-route structure, rather than independently from coordinates and
demands, so that the original route violates the added requirement and a
solver cannot succeed merely by ignoring it. Where relevant, the construction
also excludes a cost-preserving route reversal as a trivial solution.

Every instance stores both the sentence shown to humans and LLMs and a
machine-readable template with its grounded variables. The latter is used
only to build the post-hoc checker. The instruction asks
the solver to minimize total distance, obey hard constraints, and satisfy
soft requests when possible. Wording is fixed within each task to isolate
constraint-type and composition performance from paraphrase robustness. A
single pseudorandom generator is seeded once with 1234 at the beginning of
each dataset-construction run.

\begin{table}[t]
\caption{Principal fields in the public LLMCoSolver CVRP instances.}
\label{tab:tai-fields}
\centering
\small
\begin{tabularx}{\columnwidth}{@{}lY@{}}
\toprule
Field & Contents \\
\midrule
\texttt{num\_nodes} & Number of nodes, including the depot. \\
\texttt{vehicle\_capacity} & Homogeneous vehicle capacity $Q$. \\
\texttt{instruction} & Natural-language statement of unique visits,
capacity, Euclidean-distance minimization, and output format. \\
\texttt{input} & Node IDs, coordinates, demands, and the two nearest
neighbors and distances used as LLM heuristic features. \\
\texttt{output} & LKH-3 routes and total distance. \\
\bottomrule
\end{tabularx}
\end{table}

Labels above the braced positional slots state the value inserted at run time
and are not part of the recipe string. The braced token itself is reproduced
verbatim from the released recipe; one slot may expand to a comma-separated
list, as in Vehicle and Order.

\begin{constrainttemplatepair}
\paragraph{Road.}
The template is:
\begin{constrainttemplatebox}
The road between Node(*@\slotword{0}{node ID}@*)and Node(*@\slotword{1}{node ID}@*)is closed.
\end{constrainttemplatebox}
\end{constrainttemplatepair}
We first sample a reference route containing at least two customers and then
sample one adjacent customer--customer pair on that route. Depot-adjacent
edges are excluded. The selected undirected edge is therefore used by the
original route, which must change to satisfy the closure.

\begin{constrainttemplatepair}
\paragraph{Time.}
The template is:
\begin{constrainttemplatebox}
Node(*@\slotword{0}{node ID}@*)must be served between time(*@\slotword{1}{start time}@*)and(*@\slot{2}{end time}@*). However, arriving early and waiting is not allowed.
\end{constrainttemplatebox}
\end{constrainttemplatepair}
Vehicles depart at time zero, travel time from $i$ to $j$ is
$\lceil0.1c_{ij}\rceil$, service time is zero, and all unconstrained nodes
have the unrestricted window $[0,\infty)$. Waiting after an early arrival is
not allowed, so the arrival itself must lie in the stated interval.

Among routes with at least three customers, we choose the route with maximum
total travel time under this integer time model and split its customer sequence
as evenly as possible into nonempty First, Middle, and Last thirds, assigning
remainders to earlier thirds. Let
$l$ be the reference arrival time of the Middle-third customer farthest from
the depot. The window is $[l,l+w]$, with integer $w$ sampled uniformly from
$\{18,19,20\}$. This range is obtained by scaling RouteFinder's normalized
time-window-length range $[0.18,0.20]$~\citep{berto2025routefinder} by 100 to
match the integer time units above. Its target is sampled from the First third
and displayed bounds are rounded to integers. Using a later arrival band for
an early-route customer makes both the reference order and its reversal
nontrivial, while the no-wait policy prevents an arbitrarily early arrival
from satisfying the constraint.

\begin{constrainttemplatepair}
\paragraph{Vehicle.}
The template is:
\begin{constrainttemplatebox}
Nodes(*@\slotword{0}{2 node IDs}@*)must be served by Vehicle ID(*@\slot{1}{vehicle ID}@*).
\end{constrainttemplatebox}
\end{constrainttemplatepair}
We select two distinct reference routes when possible and take the
highest-demand customer from each. If their combined demand exceeds $Q$, the
largest selected demand is repeatedly replaced by the next-largest demand on
the same route until the pair fits. The designated vehicle is sampled
uniformly from the reference routes. Selecting high-demand customers from
different routes tends to require genuine redistribution rather than a
cost-free relabeling.

\begin{constrainttemplatepair}
\paragraph{Order.}
The template is:
\begin{constrainttemplatebox}
Nodes(*@\slotword{0}{2 prerequisite node IDs}@*)must be served before Nodes(*@\slotword{1}{subject node ID}@*)by the same vehicle, but not necessarily immediately before.
\end{constrainttemplatebox}
\end{constrainttemplatepair}
Two prerequisite nodes must precede one subject node. On every eligible
reference route, the subject is restricted to an interior position and the
two prerequisites include at least one customer from each side of it. This
``sandwich'' makes both the original route and its reversal violate the
requirement. Candidates whose total demand exceeds $Q$ are removed.

Among the remaining candidates, let $c_{\mathrm{left}}$ and
$c_{\mathrm{right}}$ be the centroids of prerequisites selected before and
after the subject in the reference order, and let $c_{\mathrm{after}}$ denote
the subject centroid. We maximize
\begin{equation}
\operatorname{dist}(c_{\mathrm{left}},c_{\mathrm{right}})
-\lambda\operatorname{dist}\!\left(
c_{\mathrm{after}},\overline{c_{\mathrm{left}}c_{\mathrm{right}}}
\right), \qquad \lambda=1,
\label{eq:order-score}
\end{equation}
where the second distance is to the line segment. We draw 64 random candidates
per eligible route and retain the global maximum. The score separates the
prerequisite groups while placing the subject near their connecting segment,
making low-cost routes likely to encounter the subject before all
prerequisites have been served.

\subsection{Pairwise Constraint Datasets}

The six pairwise datasets contain every pair among Road, Time, Vehicle, and
Order. Each contains the same 300 base instances. We join the corresponding
primary datasets by instance identifier. The
two natural-language conditions and their machine-readable records are
concatenated in task order, with identifiers reassigned from zero. A Road +
Time example is:
\begin{lstlisting}[style=constraint]
- The road between Node 78 and Node 37 is closed.
- Node 14 must be served between time 115 and 134. However, arriving early and waiting is not allowed.
\end{lstlisting}
Grounded targets and parameters are inherited unchanged from the two primary
datasets and are not adjusted to amplify their interaction. This construction
tests whether independently generated constraints can be satisfied jointly;
interactions may nevertheless arise naturally on the shared instance. The
single-constraint reference routes are not merged. Pairwise references are
recomputed with both constraints as described in
Section~\ref{sec:reference-solvers}.

For \emph{separate generation}, each bullet is sent in an independent LLM
request and its rule becomes a branch applied in source order. For
\emph{joint generation}, the bullet texts are joined with spaces and sent in
one request, producing one combined rule. Prompts, DSL semantics, validation,
repair, and execution are otherwise identical. With one condition the two
forms coincide.

\subsection{Additional Constraint Datasets}

The breadth evaluation covers twelve constraint types beyond the four
primary families. All use the same 50-instance subset of the base data. We
stratify by customer count into $[10,20]$, $[21,40]$, $[41,60]$,
$[61,80]$, and $[81,100]$. Proportional allocation with the largest-remainder
method yields 4, 11, 13, 11, and 11 instances; within each stratum we take the
first instances in source order. Because subsampling precedes constraint
generation, every task uses identical numerical CVRP instances.

The common protocol and seed 1234 from Section~A.1 apply. No candidate was
rejected, so every dataset contains 50 instances. Time-dependent tasks share
the Time model: departure at zero, travel time $\lceil0.1c_{ij}\rceil$, zero
service time, no waiting, and default window $[0,\infty)$. These datasets have
no constraint-specific reference solutions, so only feasibility is reported.

\begin{constrainttemplatepair}
\paragraph{First delivery.}\mbox{}\par
\begin{constrainttemplatebox}
Node(*@\slotword{0}{node ID}@*)must be the first customer served on the route of the vehicle assigned to it. The depot is not counted as a customer.
\end{constrainttemplatebox}
\end{constrainttemplatepair}
Among all customers that are neither first nor last on their reference route,
we choose the one farthest from the depot, breaking ties by node ID. Excluding
both endpoints makes the reference and its reversal violate the constraint;
the distance choice discourages a trivial low-cost reorder.

\begin{constrainttemplatepair}
\paragraph{Visit position.}\mbox{}\par
\begin{constrainttemplatebox}
Node(*@\slotword{0}{node ID}@*)must be the(*@\slot{1}{position}@*)-th customer served on the route of its assigned vehicle.
\end{constrainttemplatebox}
\end{constrainttemplatepair}
For a customer at reference position $p$ on a $k$-customer route, eligible
positions are $\{1,\ldots,k\}\setminus\{p,k-p+1\}$. We uniformly sample a
customer with a nonempty set and then an eligible position, excluding both
the reference and reversed positions.

\begin{constrainttemplatepair}
\paragraph{Arrival capacity reserve.}\mbox{}\par
\begin{constrainttemplatebox}
Node(*@\slotword{0}{node ID}@*)must be served only when the vehicle's remaining capacity is at least(*@\slot{1}{capacity threshold}@*).
\end{constrainttemplatebox}
\end{constrainttemplatepair}
Let $(c_1,\ldots,c_k)$ be the customer sequence of the longest reference
route, where $c_i$ is the customer node at position $i$ and $k$ is the number
of customers on that route. Choose a customer uniformly from positions
$c_{\lfloor k/2\rfloor+1},\ldots,c_{k-1}$. If $L$ is demand served before
that customer and $C_r=Q-L$, sample the threshold uniformly from the integers
in $[C_r+1,C_r+\lfloor\bar d\rfloor]$. The reference necessarily violates
the threshold, but the extra reserve is limited to roughly one average
delivery.

\begin{constrainttemplatepair}
\paragraph{Heavy-delivery limit.}\mbox{}\par
\begin{constrainttemplatebox}
Each vehicle may serve at most(*@\slotword{1}{customer limit}@*)customer nodes whose demand is at least(*@\slot{0}{demand threshold}@*).
\end{constrainttemplatebox}
\end{constrainttemplatepair}
Sample limit $Y$ uniformly from 1--3, reducing the upper bound to one less
than the number of distinct demands if that number is at most three. Starting
from the maximum demand, decrease threshold $X$ until some reference route
contains at least $Y+1$ customers with demand at least $X$. This is the
largest threshold that guarantees a violation and avoids labeling most
customers as heavy deliveries.

\begin{constrainttemplatepair}
\paragraph{Minimum service interval.}\mbox{}\par
\begin{constrainttemplatebox}
The same vehicle must begin service at the next customer at least(*@\slotword{1}{minimum interval}@*)time units after completing service at Node(*@\slot{0}{node ID}@*), without waiting.
\end{constrainttemplatebox}
\end{constrainttemplatepair}
Among reference customers followed by another customer, choose the one
farthest from the depot. If $t_1$ is travel time to its reference successor
and $t_2$ is travel time to the depot, set the displayed interval to the
rounded midpoint $g=(t_1+t_2)/2$. In the released 50 instances, the rounded
value satisfies $t_1<g\leq t_2$, so the reference successor is prohibited
while returning to the depot remains available.

\begin{constrainttemplatepair}
\paragraph{Depot closure.}\mbox{}\par
\begin{constrainttemplatebox}
The depot is out of service during the period from time(*@\slotword{0}{start time}@*)and(*@\slot{1}{end time}@*). However, arriving early and waiting is not allowed.
\end{constrainttemplatebox}
\end{constrainttemplatepair}
Sample a reference route and set closure start to its depot-return time. The
closure duration is the rounded mean travel time over all node pairs. Thus the
selected route violates the interval, whose width is approximately one
typical move rather than an arbitrarily narrow or broad period.

\begin{constrainttemplatepair}
\paragraph{Small-delivery assignment.}\mbox{}\par
\begin{constrainttemplatebox}
Vehicle ID(*@\slotword{0}{vehicle ID}@*)must serve all and only customer nodes whose demand is at most(*@\slot{1}{demand threshold}@*).
\end{constrainttemplatebox}
\end{constrainttemplatepair}
Let $N_v=\lfloor\sum_{i=1}^{n} d_i/Q\rfloor$ and sample a vehicle ID from
$\{0,\ldots,N_v-1\}$. Let $d_{\max}$ be the largest threshold for which the
total demand of all customers at or below it fits in one vehicle, and let
$d_{\min}=\min_{i\in V\setminus\{0\}}d_i$ be the minimum customer demand.
We sample the threshold from $[d_{\min},d_{\max}]$ and reject the sample if
the designated reference route already equals the resulting customer set.

\begin{constrainttemplatepair}
\paragraph{Return deadline.}\mbox{}\par
\begin{constrainttemplatebox}
Vehicle ID(*@\slotword{0}{vehicle ID}@*)must arrive at the depot by time(*@\slot{1}{deadline}@*).
\end{constrainttemplatebox}
\end{constrainttemplatepair}
Among reference routes with at least two customers, choose the earliest to
return and set its deadline one time unit earlier. This creates the smallest
strict violation at the earliest absolute return time while avoiding a
degenerate zero-customer route.

\begin{constrainttemplatepair}
\paragraph{Vehicle workload limit.}\mbox{}\par
\begin{constrainttemplatebox}
Vehicle ID(*@\slotword{0}{vehicle ID}@*)may visit at most(*@\slotword{1}{customer limit}@*)customer nodes.
\end{constrainttemplatebox}
\end{constrainttemplatepair}
Among routes with at least two customers, select the one with the fewest and
set its limit to one less than that count. This yields the smallest positive
absolute limit and a one-customer violation.

\begin{constrainttemplatepair}
\paragraph{Consecutive deliveries.}\mbox{}\par
\begin{constrainttemplatebox}
Nodes(*@\slot{0}{node ID}@*),(*@\slot{1}{node ID}@*), and(*@\slotword{2}{node ID}@*)must be served consecutively, in that order, by the same vehicle.
\end{constrainttemplatebox}
\end{constrainttemplatepair}
Sample a reference route with at least three customers and then three of its
customers. Uniformly select a permutation different from their reference
relative order. Same-route sampling ensures the combined demand fits $Q$;
the new requirement generally changes both order and adjacency.

\begin{constrainttemplatepair}
\paragraph{Customer incompatibility.}\mbox{}\par
\begin{constrainttemplatebox}
Nodes(*@\slotword{0}{node ID}@*)and(*@\slotword{1}{node ID}@*)must be served by different vehicles.
\end{constrainttemplatebox}
\end{constrainttemplatepair}
Sample one adjacent customer pair from a reference route with at least two
customers. The reference violates the assignment, and adjacency tends to
select geometrically compatible customers whose separation is costly.

\begin{constrainttemplatepair}
\paragraph{Restricted-area transition.}\mbox{}\par
\begin{constrainttemplatebox}
Do not travel directly from Node(*@\slotword{0}{source node ID}@*)to any customer node with an x-coordinate between(*@\slotword{1}{lower x-bound}@*)and(*@\slot{2}{upper x-bound}@*).
\end{constrainttemplatebox}
\end{constrainttemplatepair}
Sample an interior reference customer as the source. If its successor has
$x$-coordinate $x_c$ and customer extrema are $x_{\min}$ and $x_{\max}$,
the rounded bounds are $(x_{\min}+x_c)/2$ and $(x_c+x_{\max})/2$. The
interval contains the reference successor, while excluding route endpoints
prevents a depot successor in either the original route or its reversal.
Returning directly to the depot remains permitted.

\subsection{Reference Solvers and Evaluation}
\label{sec:reference-solvers}

\subsubsection{Independent post-hoc checker}

Every method is evaluated by the same checker, which receives only the
returned route sequence and is isolated from rule generation, Rule Executor
implementation, decoding, and rollout selection. A route is feasible only if
it starts and ends at the depot, respects capacity, visits every customer
exactly once, and satisfies every grounded additional constraint. Empty or
unparseable outputs are infeasible. The objective is the sum of unrounded
Euclidean edge lengths over all routes, irrespective of any integer costs
used internally by a reference solver.

Road checks both directions of the closed edge. Time reconstructs arrival
times under the no-wait model. Vehicle IDs are zero-based positions in the
returned route list, and a missing designated position is a violation. Order
requires all referenced nodes on one route and every prerequisite position
to precede the subject. Pairwise constraints are dispatched by their task
prefix and combined by conjunction. The additional tasks
use analogous task-specific checks derived from their machine-readable
records.

If DSL parsing, schema validation, and both repair attempts all fail,
LangNCO solves the instance without an additional rule; the returned route is
still checked against the requested constraint. Runtime exceptions and
decoding deadlocks are recorded as no solution. No failure is removed from
the feasibility denominator.

\subsubsection{Road references}

Road references use LKH-3~\citep{helsgaun2017lkh3} with an explicit integer
distance matrix. Each closed-edge cost is set to ten times the largest edge
length in the instance. We run ten trials with seed zero and accept a route
only after the independent checker verifies the base CVRP and road closure.

\subsubsection{Time, Vehicle, and Order references}

These references use the OR-Tools Routing Solver~\citep{ortools} with
\texttt{PARALLEL\_CHEAPEST\_INSERTION},
\texttt{GUIDED\_LOCAL\_SEARCH}, and a 600-second limit per instance. Arc
costs are ceiling-rounded Euclidean distances; reported objectives are
recomputed from unrounded distances. The vehicle count is
$\min\{n,\lceil\sum_i d_i/Q\rceil+2\}$ and is increased if needed to include
a designated vehicle ID.

Time uses a time dimension with zero slack, Vehicle fixes the routing solver's
vehicle variable, and Order combines a visit-position dimension with
same-vehicle equalities. To improve initial-solution construction, only
non-target customers may be dropped with penalty $10^5$. Target nodes remain
mandatory, and any final route with a dropped customer is rejected. Since
coordinates lie in $[0,1000]^2$, one ceiling-rounded edge costs at most 1415
and a simple insertion changes two edges by at most about 2830; the penalty
therefore strongly favors reinsertion. It is a search aid, not a proof that
every incomplete route costs more than every complete one. All accepted
primary and pairwise references visit every customer.

Empty internal routes can change vehicle positions after compression. We
therefore retain internal vehicle--route pairs, remove empty routes, construct
an explicit map to compressed output positions, and transform grounded
vehicle IDs before checking.

\subsubsection{Pairwise and additional references}

Pairwise references impose both constraints in one OR-Tools model; closed
edges are removed from successor domains. We compare six first-solution
strategies and select \texttt{GLOBAL\_CHEAPEST\_ARC}, which has the greatest
aggregate reference coverage (Table~\ref{tab:ortools-strategy}). Other
settings match the single-constraint OR-Tools runs. A solution is retained
only if it passes both checks. Failure within 600 seconds is not an
infeasibility certificate. No constrained reference solver is constructed
for the twelve additional tasks, whose purpose is breadth rather than
objective-gap evaluation.

\subsubsection{Metrics}

Feasibility is the fraction of all test instances whose output passes the
checker. For a feasible output with objective $z$ and an available constrained
reference objective $z^*$,
\begin{equation}
\operatorname{gap}(z,z^*)=\frac{z-z^*}{z^*}\times100\%.
\label{eq:gap}
\end{equation}
Primary-task gaps are averaged over feasible method outputs; all 300 primary
instances have references. Pairwise feasibility counts always use all 300
instances for each method. Pairwise gaps use only the intersection on which
that LangNCO variant and OR-Tools both return feasible solutions. Additional
tasks report feasibility over all 50 instances and no reference gap. Reference
solutions are time-limited, near-optimal baselines rather than
optimality-certified ground truth.

\section{LangNCO Implementation Details}
\label{sec:implementation}

\subsection{DSL Schema, Validation, and Compilation}

\subsubsection{Typed guarded-rule representation}

The LangNCO DSL is a typed guarded mask-effect intermediate representation.
It separates four semantic components, summarized in
Table~\ref{tab:dsl-components}, rather than asking the LLM to generate
executable tensor code.

\begin{table*}[t]
\caption{Semantic components of the guarded-rule representation.}
\label{tab:dsl-components}
\centering
\small
\begin{tabularx}{\textwidth}{@{}lY@{}}
\toprule
Component & Abstract role \\
\midrule
Target & Selects the candidate actions affected by a constraint. \\
State guard & Determines whether the constraint is active for a decoder
state. \\
Candidate predicate & Evaluates the constraint separately for each candidate
action. \\
Effect & Removes matched actions or restricts the feasible set to matched
actions. \\
\bottomrule
\end{tabularx}
\end{table*}

Targets may identify fixed nodes, structural sets such as the depot or all
customers, or sets selected from node attributes. Expressions combine typed
Boolean, comparison, arithmetic, and visit-history aggregation primitives.
Accessors expose standard CVRP state, cumulative or vehicle-specific visit
history, node attributes, and explicitly declared auxiliary observations.
The latter may be scalar, node indexed, vector valued, or pairwise. In these
experiments, elapsed time and travel times are auxiliary observations visible
to the Rule Executor but not added to the frozen NCO policy input.

For constraint $c$, the matched-action set is
\begin{equation}
M_c(s_t)=\left\{a\in T_c\,\middle|\,
g_c(s_t)\land h_c(s_t,a)\right\},
\label{eq:dsl-match}
\end{equation}
where $T_c$ is the target set, $g_c(s_t)\in\{0,1\}$ is a state-level guard,
and $h_c(s_t,a)\in\{0,1\}$ is a candidate-level predicate. With batch size
$B$ and $N$ nodes, state guards are evaluated in a row context and return
$B$ Booleans, whereas target predicates and candidate guards are evaluated
in a node context and return $B\times N$ Booleans. Static context checks
reject candidate-indexed access where only a row value is valid. This
separation delegates broadcasting, tensor shapes, types, and device placement
to the compiler rather than the LLM.

\subsubsection{Structured generation, validation, and compilation}

The generation protocol exposes a typed catalog of available observations,
target constructors, expression families, and nonexpansive effects. It asks
for one structured rule representation without executable code or explanatory
text. For multiple requirements, accepted constraints are combined in source
order, keeping language interpretation separate from deterministic execution.

After parsing and normalization, the rule is validated against a
machine-readable schema. The same compiler then performs a dry compilation,
rejecting unsupported operations, malformed target sets, missing arguments,
and context-invalid access. Dry compilation establishes structural
conformance and static translatability, assuming run-time fields have their
declared shapes; it does not execute a route or establish semantic agreement
with the sentence.

On a parse, schema, or compile error, the prior output and concrete diagnostic
are returned to the LLM for at most two bounded repairs. If all three attempts
fail, the instance is solved with only the base CVRP mask. This fallback
preserves pipeline execution but does not satisfy the omitted requirement by
default. The protocol describes the information and guarantees supplied to
the generator; exact prompt wording and serialization details are not needed
for the execution semantics analyzed here.

For a valid constraint, the compiler evaluates $g_c$ as a row mask in
$\{0,1\}^{B}$ and represents $T_c$ and $h_c$ as target and candidate
masks $T_c,h_c\in\{0,1\}^{B\times N}$. It constructs the Boolean match mask
$K_c\in\{0,1\}^{B\times N}$ whose entries, for $b=1,\ldots,B$ and $a\in V$,
are
\begin{equation}
K_c[b,a]=T_c[b,a]\land g_c[b]\land h_c[b,a].
\label{eq:compiled-match}
\end{equation}
Thus, $K_c[b,a]=1$ if and only if $a\in M_c(s_t^{(b)})$. As a minor
executor artifact, a
\texttt{force\_only} update is evaluated only on rows selected by $g_c$;
unselected rows pass through unchanged.
Fixed row values are broadcast over candidates, while \texttt{\$n} access is
vectorized. All generated tensors use explicit types and the input mask's
device.

DSL thresholds use raw units. If RouteFinder stores normalized coordinate and
demand values $x_i^{\mathrm{norm}}$ and $d_i^{\mathrm{norm}}$, the compiler
recovers $x_i^{\mathrm{raw}}=1000x_i^{\mathrm{norm}}$ and
$d_i^{\mathrm{raw}}=Qd_i^{\mathrm{norm}}$. Coordinate predicates use the
canonical, unaugmented coordinates so geometric augmentation cannot change a
constraint's meaning. A rule needing unavailable auxiliary state or
vehicle-specific history is inactive and leaves the input mask unchanged.

For an active constraint, starting with the base feasible set $\mathcal A$ and
apart from the exhausted-target safeguard described below, source-ordered
constraints apply
\begin{align}
\mathsf{block}:\quad
\mathcal A&\leftarrow\mathcal A\setminus M_c(s_t),\nonumber\\
\mathsf{force\_only}:\quad
\mathcal A&\leftarrow\mathcal A\cap M_c(s_t).
\label{eq:dsl-effects-supp}
\end{align}
Both effects are intersections or removals from the current mask; no later
rule can re-enable a base- or earlier-rule-prohibited action.

\subsection{Rule Executor and Deadlock Handling}

At each step, the Rule Executor takes RouteFinder's standard-CVRP base mask
and applies every compiled constraint and branch in order. Let
$m_{t,q}^{\mathrm{rule}}(a)\in\{0,1\}$ denote the retain mask contributed by
update $q$. The final mask satisfies
\begin{equation}
\begin{aligned}
m_t(a)&=m_t^{\mathrm{base}}(a)\land
\bigwedge_q m_{t,q}^{\mathrm{rule}}(a),\\
m_t(a)&\leq m_t^{\mathrm{base}}(a),\qquad a\in V.
\end{aligned}
\end{equation}
It is shared by greedy and sampling decoders; masked logits are set to
$-\infty$.

A limited safeguard skips \texttt{force\_only} when its target contains
neither the depot nor any unvisited customer. It does not skip a temporarily
unavailable but unvisited customer, which may become available after a depot
return. The safeguard keeps execution defined after an exhausted target but
does not certify the skipped constraint.

A rollout is deadlocked if customers remain while no customer or depot action
is executable. A depot state with no visitable customer is also deadlocked,
because repeatedly selecting the depot cannot change the state. Likewise, a
POMO start that externally forces an action prohibited by the executed mask
is marked deadlocked. Before masked softmax, an all-false row is assigned a
temporary depot-only numerical sink so batch shapes remain valid; the rollout
is terminal with reward $-\infty$, so this sink cannot recover or select a
solution. If every rollout deadlocks, the instance returns no solution. There
is no backtracking, mask relaxation, or rule regeneration. Vectorized
execution of $R$ constraints over $|V|$ actions costs $O(R|V|)$ per step.

\subsection{Critical-Node Extraction and Critical-Offset Decoding}

Hard masking establishes only immediate legality. A legal action can still
consume capacity required by an assigned customer, miss a future time window,
or close a route before an ordering prerequisite is completed. Critical-Offset
Decoding uses DSL-referenced nodes as a low-cost completion-oriented rollout
prior instead of solving a constrained continuation for every candidate.

For a rule set $r$ and constraint $c\in r$, let $N_{\mathrm{tar}}(c)$ be
fixed IDs resolved from its target and $N_{\mathrm{cond}}(c)$ fixed IDs
resolved from its guards. We call them subject and condition nodes and define
\begin{align}
\mathcal S_r&=\bigcup_{c\in r}N_{\mathrm{tar}}(c),\nonumber\\
\mathcal G_r&=\bigcup_{c\in r}N_{\mathrm{cond}}(c),\nonumber\\
\mathcal C_r&=(\mathcal S_r\cup\mathcal G_r)\setminus\{0\}.
\label{eq:critical-set-supp}
\end{align}
The depot is excluded to avoid encouraging premature route termination.
Extraction is deterministic, scope-aware, and shared across all rollouts of
an instance. It recursively collects static targets and fixed IDs in
current-node comparisons and visit aggregations. State values such as time,
capacity, and vehicle ID add no condition node. Attribute-defined
\texttt{where} targets and \texttt{all\_customers} are not expanded into the
lookup table in the current implementation, although their hard rules still
execute normally.

At step $t$, only legal critical nodes are active:
\begin{equation}
\mathcal A_t=\{a:m_t(a)=1\},\qquad
\mathcal C_t=\mathcal C_r\cap\mathcal A_t.
\end{equation}
For masked-policy probability $p_t(a)$ and strength $\alpha\in[0,1]$, define
the unnormalized weight
\begin{equation}
q_t^{(\alpha)}(a)=
\begin{cases}
(1-\alpha)p_t(a)+\alpha,&a\in\mathcal C_t,\\
p_t(a),&\text{otherwise}.
\end{cases}
\label{eq:offset-supp}
\end{equation}
Greedy decoding maximizes $q_t^{(\alpha)}$; sampling normalizes it over
$\mathcal A_t$. At $\alpha=0$ this is exactly the original masked policy. At
$\alpha=1$, legal critical nodes each receive unnormalized weight one, but
noncritical legal actions retain positive policy weight. An offset never
revives a masked action.

For every POMO start and each of eight geometric augmentations, the total
sampling budget is 100. Ninety rollouts use $\alpha=0$ and two each use
\begin{equation}
\Lambda=\{0.2,0.4,0.6,0.8,1.0\}.
\end{equation}
This fixed 90:10 allocation and strength set were selected heuristically, not
tuned on a validation set. Complete rollouts are ranked only by travel cost;
incomplete and deadlocked rollouts are discarded. The independent checker is
not used to filter or rank candidates. Critical lookup and weighting add
$O(|V|)$ time per step and $O(B|V|)$ lookup storage, without increasing the
number of policy evaluations relative to sampling with 100 rollouts per POMO
start and augmentation.

\subsection{Frozen RouteFinder and Input Audit}

We train RouteFinder~\citep{berto2025routefinder} only on standard CVRP and
freeze the same parameters for all experiments. There is no fine-tuning,
gradient update, optimizer step, or online adaptation on augmented instances.
Let $x_i^{\mathrm{norm}}=x_i/1000$ and
$d_i^{\mathrm{norm}}=d_i/Q$ be the coordinate and linehaul-demand inputs to
RouteFinder. In the CVRP-specific implementation, the initial embeddings and
step context are
\begin{align}
e_0^{(0)}&=W_{\mathrm{dep}}x_0^{\mathrm{norm}},\\
e_i^{(0)}&=W_{\mathrm{cust}}
  [x_i^{\mathrm{norm}},d_i^{\mathrm{norm}}],
  \quad i\in V\setminus\{0\},\\
h_t^{\mathrm{ctx}}&=W_{\mathrm{ctx}}[e_{v_t},1-u_t/Q].
\end{align}
Here the three $W$ matrices are learned linear projections, $e_i$ is the
encoder output obtained from $e_i^{(0)}$, $v_t$ is the current node, and $u_t$
is the raw linehaul demand already served on the current route; hence
$1-u_t/Q$ is the normalized available load. Thus the learned initial embedding
receives only depot/customer coordinates and customer linehaul demand, while
the learned step context receives only the current-node encoder embedding and
available load.
Unlike the multi-variant RouteFinder embedding, this CVRP-specific embedding
has no input channels for time windows, service times, backhaul demand,
open-route flags, distance limits, current time, or remaining distance.

\begin{table*}[t]
\caption{Information-flow audit. Constraint-specific state affects the
frozen policy only through the nonexpansive action mask and legal-action
offset.}
\label{tab:input-audit}
\centering
\small
\begin{tabularx}{\textwidth}{@{}lYY@{}}
\toprule
Path & Used & Not used \\
\midrule
Learned initial embedding & Depot/customer coordinates and customer demand &
Time windows, service time, backhaul demand, open-route flag, distance limit,
natural language, or DSL tokens \\
Learned step context & Current-node embedding and available load & Current
time, vehicle index, full visit sequence, or remaining distance \\
Decoder interface & Combined base/rule mask, critical-node lookup, and
$\alpha$ & Natural-language or DSL tokens and raw values referenced by guards \\
Environment and Rule Executor & Visit state, time/travel state, vehicle ID,
capacity, and other state needed to evaluate a mask & Any learned-parameter
update \\
\bottomrule
\end{tabularx}
\end{table*}

The environment may retain extra state to execute a hard rule, but that state
is converted to a Boolean mask before affecting the policy. Critical-Offset
Decoding similarly converts statically extracted node IDs to a lookup and
changes weights only among already legal actions. With an empty rule set, the
overlay returns the base mask and $\mathcal C_t=\varnothing$, so for every
$\alpha$,
\begin{equation}
r=\varnothing\Longrightarrow
m_t=m_t^{\mathrm{base}},\qquad q_t^{(\alpha)}=p_t.
\end{equation}
Thus, with the same checkpoint, decoding settings, seed, and budget, empty-rule
LangNCO follows the ordinary RouteFinder code path and sampling distribution.

\section{Experimental Settings and Reproducibility}
\label{sec:reproducibility}

\subsection{Models and Baselines}

We generate rules with locally served Gemma4 and with GPT-5 through Azure
OpenAI. DSL-versus-Python comparisons hold the LLM, sentence, frozen NCO
solver, and decoding budget fixed while using the representation-specific
generation and repair protocols reported below. Table~\ref{tab:llm-settings}
records the generation settings.

\begin{table*}[t]
\caption{LLM rule-generation settings. ``Default'' means the request omitted
that option. Context length is shared by prompt and output; maximum completion
is the request-level output cap.}
\label{tab:llm-settings}
\centering
\small
\begin{tabularx}{\textwidth}{@{}lYY@{}}
\toprule
Setting & Gemma4 & GPT-5 \\
\midrule
Identifier & \texttt{gemma4:31b-it-q8\_0} & Azure deployment \texttt{gpt-5} \\
Runtime/API & Ollama & Azure OpenAI API, \texttt{2025-01-01-preview} \\
Quantization & 8 bit & Managed API (not applicable) \\
Temperature & 0.0 & Default \\
Top-$p$/top-$k$ & Ollama defaults & Defaults \\
Context length & 8,192 tokens
(\texttt{OLLAMA\_CONTEXT\_LENGTH=8192}) & Managed by API; not configured \\
Maximum completion & Not set in request & 8,192 tokens; 16,384 only for Primary
Order Python generation \\
Reasoning effort & Not applicable & \texttt{medium} \\
Generation seed & 0 & No reproducible fixed seed \\
DSL repair & Up to two validation-feedback attempts & Same \\
Python repair & None & None \\
\bottomrule
\end{tabularx}
\end{table*}

The implementation passed seed zero to GPT-5, but the API did not guarantee
determinism, so we do not treat it as a reproducible seed. GPT-5 requests omit
temperature because the deployment does not accept an arbitrary value. Only
Primary Order free-form Python required 16,384 completion tokens after an
8,192-token context-length error; no DSL condition used the larger limit.

\begin{table*}[t]
\caption{Methods and inference settings. Every RouteFinder-based row uses the
same frozen standard-CVRP policy.}
\label{tab:method-settings}
\centering
\small
\begin{tabularx}{\textwidth}{@{}lYY@{}}
\toprule
Method/component & Model or implementation & Inference setting \\
\midrule
LangNCO & Frozen RouteFinder & Critical-Offset Decoding; eight dihedral
augmentations; all POMO starts; 100 samples per start and augmentation;
decoder temperature 1.0; 90 ordinary and ten offset-guided samples. \\
\addlinespace[2pt]
Direct GPT-5 & Azure OpenAI \texttt{gpt-5} & One-shot prompt; one response per
instance; 128,000 maximum completion tokens; \texttt{medium} reasoning;
default temperature; no retry; regex and JSON route parsing. \\
\addlinespace[2pt]
LLMCoSolver & Retrained Qwen2.5-7B-based model & Two-nearest-neighbor
augmentation; best of eight; temperature 0.7; top-$p$ 0.9; 3,000 maximum new
tokens; selection by base-CVRP reward. \\
\addlinespace[2pt]
NCO w/o intervention & Same frozen RouteFinder & No rule; augmented
multi-start greedy decoding; eight augmentations and all POMO starts. \\
\addlinespace[2pt]
LKH-3 reference & Road wrapper & \texttt{RUNS=10}; seed zero; closed edge
cost ten times the maximum instance edge. \\
\addlinespace[2pt]
OR-Tools reference & Routing Solver & PCI for Time/Vehicle/Order and GCA for
Pairwise; guided local search; 600 seconds per instance. \\
\bottomrule
\end{tabularx}
\end{table*}

One hundred samples means 100 rollouts inside every augmentation--POMO-start
combination, not 100 trajectories total per instance. The samples and the two
rollouts at each nonzero $\alpha$ are internal candidates within one algorithm
invocation, not repeated experimental runs.

Direct GPT-5 receives one feasible route generated by a randomized heuristic
as its one-shot example. LLMCoSolver retains its Alpaca-style scaffold and
two-nearest-neighbor input, with only the new condition and required problem
definition appended.

We retrain LLMCoSolver following its published procedure: one epoch of CVRP
supervised fine-tuning with per-device batch size four, followed by the
published reinforcement-learning stage initialized from the final SFT
checkpoint, and then weight merging. Other settings use the released
defaults. We correct one response-delimiter issue in the public SFT collator:
Qwen tokenizes \texttt{\#\#\# Response:} differently at the beginning of a
string and after whitespace. We tokenize it after a whitespace-ending dummy
prefix so prompt labels are masked and loss is computed only on the response.
Training data, architecture, and hyperparameters are otherwise unchanged.
The retrained baseline obtains 100\% feasibility and a 4.91\% gap on
unconstrained CVRP, close to the reported 100\% and 4.53\%.

\subsection{Randomness and Number of Runs}

\begin{table*}[t]
\caption{Randomness controls and deterministic selection rules.}
\label{tab:seeds}
\centering
\small
\begin{tabularx}{\textwidth}{@{}lY@{}}
\toprule
Stage & Seed or selection rule \\
\midrule
Primary and Additional generation & 1234 for every task. \\
Pairwise construction & Join the two primary instances by index; no new
sampling. \\
Additional subset & Proportional size strata; first source instances within
each stratum. \\
Gemma4 rule generation & Seed 0 and temperature 0.0. \\
GPT-5 rule generation / Direct GPT-5 & No reproducible fixed seed. \\
LangNCO NCO sampling & Seed $i$ for instance $i$. \\
NCO w/o intervention & Seed 0; decoding is greedy. \\
LKH-3 & Seed 0. \\
OR-Tools and LLMCoSolver & No explicit random seed. \\
\bottomrule
\end{tabularx}
\end{table*}

Each method is invoked once per test instance. Direct GPT-5 makes one request
and does not retry API or parse failures. DSL repair attempts are internal
pipeline operations triggered only by validation failure. LLMCoSolver's eight
candidates likewise form one best-of-eight invocation. The protocol does not
estimate training-seed or API-generation variance. GPT-5 results are therefore
not exactly reproducible.

\subsection{Hardware, Software, and Timing Protocol}

\begin{table}[t]
\caption{Local execution environment.}
\label{tab:hardware}
\centering
\small
\begin{tabularx}{\columnwidth}{@{}lY@{}}
\toprule
Item & Configuration \\
\midrule
CPU & Intel Xeon w9-3575X, 44 cores / 88 threads \\
GPU & NVIDIA RTX PRO 6000 Blackwell Max-Q, 96 GB \\
Memory & 512 GiB \\
OS & Ubuntu 24.04 \\
Python & 3.12 \\
Ollama & 0.30.8 \\
LKH-3 & 3.0.8 \\
\bottomrule
\end{tabularx}
\end{table}

End-to-end latency is measured separately for each instance and decomposed
into (i) rule generation, including repair; (ii) validation and DSL-to-Python
compilation, or loading free-form code; and (iii) NCO solving. Before timing
Ollama, one dummy request excludes initial model loading. The GPU is
synchronized immediately before and after NCO timing; no separate NCO warm-up
solve is used. Compilation includes validation, code generation, and function
loading, but excludes dataset and checkpoint loading, Ollama loading, output
writing, and post-hoc evaluation.

LangNCO processes the test instances sequentially, one at a time with batch
size one; its three measured stages are also sequential within each instance.
Direct GPT-5 requests were issued in parallel, so their total wall time is not
directly comparable.
Observed API latency includes network transfer, service queueing, and
service-side inference; Ollama latency also includes local client--server
communication. The reported software versions, timing boundaries, and
execution order define the protocol used for the end-to-end latency
measurements in Table~\ref{tab:latency}.

\section{Additional Experimental Results}
\label{sec:results}

Feasibility in every table is determined by the independent checker.
Objective and gap entries are mean $\pm$ standard deviation over feasible
outputs only. Consequently, mean objectives are not directly comparable when
methods have different feasible subsets. A gap is computed between a feasible
output and its corresponding constrained reference.

\subsection{Unconstrained CVRP Sanity Check}

Table~\ref{tab:base-sanity} verifies each solver on the 300 unmodified CVRP
instances and checks that an empty LangNCO rule has no effect. The reference
objective over all instances is $9{,}662.95\pm4{,}737.80$.

\begin{table*}[t]
\caption{Unconstrained CVRP sanity check. The empty-rule result exactly
matches the frozen RouteFinder baseline.}
\label{tab:base-sanity}
\centering
\small
\begin{tabular}{@{}lrrr@{}}
\toprule
Method & Feasible & Gap [\%] & Objective \\
\midrule
LLMCoSolver & 300/300 (100.00\%) & $4.91\pm3.37$ & $10{,}227.09\pm5{,}208.21$ \\
Direct GPT-5 & 277/300 (92.33\%) & $48.76\pm82.40$ & $15{,}326.47\pm12{,}776.96$ \\
NCO w/o intervention & 300/300 (100.00\%) & $2.83\pm2.32$ & $9{,}885.24\pm4{,}750.93$ \\
LangNCO (empty rule) & 300/300 (100.00\%) & $2.83\pm2.32$ & $9{,}885.24\pm4{,}750.93$ \\
\bottomrule
\end{tabular}
\end{table*}

The retrained LLMCoSolver's 100\% feasibility and 4.91\% gap confirm that it
learned the base problem before testing unseen conditions. Direct GPT-5 fails
on 23 instances and has a 48.76\% feasible-output gap. RouteFinder and
empty-rule LangNCO have identical outputs, feasibility, and cost statistics.

\subsection{Full End-to-End and Decoding Results}

\subsubsection{End-to-end performance}

Table~\ref{tab:e2e-full} reports the complete natural-language-to-route
pipeline. ``--'' means no feasible output, so gap and objective are undefined.

\begin{table*}[t]
\caption{Full primary-task end-to-end results. Gaps and objectives summarize
only feasible outputs; each dataset has 300 instances.}
\label{tab:e2e-full}
\centering
\footnotesize
\begin{tabular}{@{}llrrr@{}}
\toprule
Dataset & Method & Feasible & Gap [\%] & Objective \\
\midrule
Road & LLMCoSolver & 109/300 (36.33\%) & $5.70\pm3.34$ & $11{,}811.25\pm5{,}577.29$ \\
Road & Direct GPT-5 & 285/300 (95.00\%) & $35.02\pm64.20$ & $13{,}281.54\pm9{,}665.37$ \\
Road & NCO w/o intervention & 112/300 (37.33\%) & $2.65\pm2.02$ & $10{,}865.53\pm5{,}121.04$ \\
Road & LangNCO-Gemma4 & 300/300 (100.00\%) & $1.63\pm1.65$ & $9{,}828.66\pm4{,}719.14$ \\
Road & LangNCO-GPT-5 & 300/300 (100.00\%) & $1.63\pm1.65$ & $9{,}828.66\pm4{,}719.14$ \\
\addlinespace
Time & LLMCoSolver & 41/300 (13.67\%) & $4.42\pm2.46$ & $11{,}325.59\pm5{,}269.24$ \\
Time & Direct GPT-5 & 239/300 (79.67\%) & $120.01\pm123.02$ & $23{,}559.14\pm17{,}924.90$ \\
Time & NCO w/o intervention & 33/300 (11.00\%) & $1.63\pm2.64$ & $11{,}104.62\pm4{,}928.38$ \\
Time & LangNCO-Gemma4 & 300/300 (100.00\%) & $1.89\pm3.00$ & $9{,}953.71\pm4{,}696.32$ \\
Time & LangNCO-GPT-5 & 300/300 (100.00\%) & $1.89\pm3.00$ & $9{,}953.71\pm4{,}696.32$ \\
\addlinespace
Vehicle & LLMCoSolver & 6/300 (2.00\%) & $0.58\pm0.49$ & $6{,}172.51\pm1{,}326.12$ \\
Vehicle & Direct GPT-5 & 286/300 (95.33\%) & $40.39\pm68.58$ & $14{,}498.03\pm11{,}035.21$ \\
Vehicle & NCO w/o intervention & 3/300 (1.00\%) & $2.56\pm1.12$ & $6{,}952.16\pm408.46$ \\
Vehicle & LangNCO-Gemma4 & 300/300 (100.00\%) & $2.53\pm2.65$ & $10{,}158.41\pm4{,}809.64$ \\
Vehicle & LangNCO-GPT-5 & 300/300 (100.00\%) & $2.68\pm2.87$ & $10{,}167.84\pm4{,}803.62$ \\
\addlinespace
Order & LLMCoSolver & 0/300 (0.00\%) & \NA & \NA \\
Order & Direct GPT-5 & 287/300 (95.67\%) & $32.04\pm57.23$ & $13{,}185.65\pm9{,}079.90$ \\
Order & NCO w/o intervention & 4/300 (1.33\%) & $1.47\pm0.93$ & $8{,}356.78\pm2{,}365.50$ \\
Order & LangNCO-Gemma4 & 300/300 (100.00\%) & $3.26\pm3.71$ & $10{,}050.50\pm4{,}753.62$ \\
Order & LangNCO-GPT-5 & 300/300 (100.00\%) & $2.97\pm3.46$ & $10{,}027.54\pm4{,}756.53$ \\
\bottomrule
\end{tabular}
\end{table*}

Reference objectives for Road, Time, Vehicle, and Order are respectively
$9{,}702.69\pm4{,}720.65$, $9{,}840.66\pm4{,}788.95$,
$9{,}967.52\pm4{,}866.50$, and $9{,}815.17\pm4{,}831.95$. Both LangNCO
variants return 300 feasible routes on every primary task with gaps of
1.63--3.26\%. NCO w/o intervention is feasible on only 1.00--37.33\%; its
occasionally small conditional gap describes the few accidental successes,
not general constraint handling. Direct GPT-5 is more feasible but has gaps
of 32.04--120.01\%. LLMCoSolver is fully feasible on standard CVRP but reaches
only 0.00--36.33\% under the unseen conditions.

\begin{table*}[t]
\caption{Per-instance LangNCO wall-clock latency (mean $\pm$ standard
deviation). End-to-end time is the sum of the three measured stages.}
\label{tab:latency}
\centering
\footnotesize
\begin{tabular}{@{}llrrrr@{}}
\toprule
Dataset & Generator & End-to-end [ms] & Rule generation [ms] & Validate + compile [ms] & NCO solve [ms] \\
\midrule
Road & Gemma4 & $16{,}398.08\pm4{,}678.31$ & $16{,}113.54\pm4{,}682.77$ & $2.84\pm0.72$ & $281.69\pm153.71$ \\
Road & GPT-5 & $10{,}714.05\pm3{,}681.74$ & $10{,}457.49\pm3{,}683.47$ & $0.65\pm0.24$ & $255.91\pm137.14$ \\
Time & Gemma4 & $44{,}983.17\pm6{,}997.72$ & $44{,}691.05\pm6{,}962.60$ & $2.81\pm0.79$ & $289.32\pm175.59$ \\
Time & GPT-5 & $39{,}842.02\pm22{,}012.27$ & $39{,}495.22\pm21{,}987.02$ & $0.74\pm0.26$ & $346.05\pm205.16$ \\
Vehicle & Gemma4 & $15{,}023.30\pm2{,}863.01$ & $14{,}743.96\pm2{,}829.72$ & $2.77\pm0.77$ & $276.57\pm156.65$ \\
Vehicle & GPT-5 & $36{,}637.52\pm22{,}892.23$ & $36{,}323.33\pm22{,}883.83$ & $0.64\pm0.21$ & $313.55\pm186.47$ \\
Order & Gemma4 & $72{,}362.71\pm28{,}394.11$ & $72{,}086.05\pm28{,}391.66$ & $2.85\pm0.78$ & $273.80\pm159.02$ \\
Order & GPT-5 & $48{,}958.89\pm23{,}017.82$ & $48{,}638.66\pm23{,}010.17$ & $0.74\pm0.70$ & $319.50\pm199.13$ \\
\bottomrule
\end{tabular}
\end{table*}

Validation and compilation remain below 3 ms in every condition; LLM rule
generation dominates end-to-end latency. These are client-observed times,
not isolated model-compute benchmarks, particularly for the remote GPT-5 API.

\subsubsection{Full decoding ablation}

This ablation replaces generated masks with independently implemented
reference masks to isolate decoding from rule-generation error. Critical nodes
are still extracted from LangNCO-Gemma4 DSL, not dataset labels. ``Noaug +
single start'' uses neither geometric augmentation nor POMO multistart;
``noaug'' uses POMO starts without augmentation. NCO greedy uses both. The
greedy controls add either one ordinary sample or one $\alpha=1$ offset-guided
rollout. Both sampling rows use exactly 100 rollouts per POMO start and
augmentation. Timing starts after five warm-up solves, which are excluded from
the reported values. Because the
implementation cannot execute greedy and sampling decoding in parallel,
NCO greedy + 1 sample runs them sequentially; its reported time is the
sum of the two decoder passes and is consequently longer than either pass
alone.

\begin{table*}[t]
\caption{Full decoding ablation across all four primary datasets. Time is
per-instance NCO decoding time in milliseconds; language generation and
compilation are excluded.}
\label{tab:decoding-ablation}
\centering
\footnotesize
\begin{tabularx}{\textwidth}{@{}lYrrrr@{}}
\toprule
Dataset & Strategy & Feasible & Gap [\%] & Objective & Time [ms] \\
\midrule
Road & NCO greedy (noaug + single start) & 300/300 (100.00\%) & $9.47\pm11.49$ & $10{,}424.60\pm4{,}756.40$ & $99.20\pm53.19$ \\
Road & NCO greedy (noaug) & 300/300 (100.00\%) & $5.23\pm5.84$ & $10{,}107.41\pm4{,}734.66$ & $113.87\pm50.90$ \\
Road & NCO greedy & 300/300 (100.00\%) & $3.19\pm3.04$ & $9{,}950.00\pm4{,}728.90$ & $117.36\pm56.35$ \\
Road & NCO greedy + 1 sample & 300/300 (100.00\%) & $2.28\pm2.38$ & $9{,}920.20\pm4{,}728.38$ & $239.61\pm106.60$ \\
Road & NCO sampling & 300/300 (100.00\%) & $1.64\pm1.68$ & $9{,}830.36\pm4{,}723.48$ & $167.88\pm99.84$ \\
Road & Critical-Offset greedy & 300/300 (100.00\%) & $3.15\pm3.03$ & $9{,}946.12\pm4{,}727.31$ & $140.88\pm65.04$ \\
Road & Critical-Offset sampling & 300/300 (100.00\%) & $1.63\pm1.65$ & $9{,}828.66\pm4{,}719.14$ & $173.00\pm101.77$ \\
\addlinespace
Time & NCO greedy (noaug + single start) & 158/300 (52.67\%) & $8.95\pm6.34$ & $12{,}147.49\pm5{,}023.99$ & $100.25\pm45.62$ \\
Time & NCO greedy (noaug) & 285/300 (95.00\%) & $7.20\pm7.36$ & $10{,}532.27\pm4{,}698.93$ & $120.82\pm55.04$ \\
Time & NCO greedy & 299/300 (99.67\%) & $4.47\pm4.98$ & $10{,}170.02\pm4{,}701.39$ & $119.71\pm53.89$ \\
Time & NCO greedy + 1 sample & 299/300 (99.67\%) & $3.83\pm4.35$ & $10{,}118.89\pm4{,}698.80$ & $241.94\pm109.93$ \\
Time & NCO sampling & 299/300 (99.67\%) & $2.19\pm3.33$ & $9{,}985.64\pm4{,}686.60$ & $178.90\pm111.28$ \\
Time & Critical-Offset greedy & 300/300 (100.00\%) & $3.52\pm3.71$ & $10{,}088.72\pm4{,}711.08$ & $134.63\pm67.56$ \\
Time & Critical-Offset sampling & 300/300 (100.00\%) & $1.89\pm3.00$ & $9{,}953.71\pm4{,}696.32$ & $183.36\pm113.30$ \\
\addlinespace
Vehicle & NCO greedy (noaug + single start) & 94/300 (31.33\%) & $13.67\pm14.19$ & $9{,}984.04\pm4{,}002.33$ & $96.43\pm44.75$ \\
Vehicle & NCO greedy (noaug) & 169/300 (56.33\%) & $10.05\pm10.60$ & $9{,}583.89\pm4{,}291.81$ & $112.17\pm50.58$ \\
Vehicle & NCO greedy & 226/300 (75.33\%) & $5.64\pm4.68$ & $9{,}554.24\pm4{,}496.91$ & $109.39\pm49.89$ \\
Vehicle & NCO greedy + 1 sample & 232/300 (77.33\%) & $5.26\pm4.69$ & $9{,}536.46\pm4{,}471.97$ & $228.63\pm101.83$ \\
Vehicle & NCO sampling & 263/300 (87.67\%) & $3.33\pm2.98$ & $9{,}699.06\pm4{,}630.74$ & $163.24\pm96.72$ \\
Vehicle & Critical-Offset greedy & 300/300 (100.00\%) & $4.56\pm4.14$ & $10{,}316.78\pm4{,}793.07$ & $126.06\pm65.33$ \\
Vehicle & Critical-Offset sampling & 300/300 (100.00\%) & $2.53\pm2.65$ & $10{,}158.41\pm4{,}809.64$ & $168.07\pm97.95$ \\
\addlinespace
Order & NCO greedy (noaug + single start) & 27/300 (9.00\%) & $9.91\pm6.46$ & $8{,}141.98\pm3{,}175.66$ & $98.35\pm44.92$ \\
Order & NCO greedy (noaug) & 134/300 (44.67\%) & $11.39\pm11.47$ & $9{,}418.80\pm4{,}154.35$ & $117.93\pm54.49$ \\
Order & NCO greedy & 225/300 (75.00\%) & $7.69\pm7.42$ & $9{,}789.76\pm4{,}658.21$ & $115.28\pm52.47$ \\
Order & NCO greedy + 1 sample & 251/300 (83.67\%) & $6.56\pm6.46$ & $9{,}938.24\pm4{,}722.22$ & $231.92\pm103.10$ \\
Order & NCO sampling & 291/300 (97.00\%) & $3.52\pm4.18$ & $9{,}866.84\pm4{,}635.28$ & $185.10\pm106.92$ \\
Order & Critical-Offset greedy & 300/300 (100.00\%) & $6.81\pm6.40$ & $10{,}325.12\pm4{,}751.61$ & $121.99\pm53.50$ \\
Order & Critical-Offset sampling & 300/300 (100.00\%) & $2.85\pm3.30$ & $10{,}018.17\pm4{,}759.80$ & $174.28\pm103.07$ \\
\bottomrule
\end{tabularx}
\end{table*}

Adding one ordinary rollout raises greedy feasibility only from 75.33\% to
77.33\% on Vehicle and from 75.00\% to 83.67\% on Order. Replacing that
rollout with one offset-guided trajectory reaches 100\% on both. At the same
budget of 100 rollouts per POMO start and augmentation, ordinary sampling
fails on 1 Time, 37 Vehicle, and 9 Order instances, while Critical-Offset
sampling reaches 100\% on all four tasks and
reduces their gaps from 2.19\%, 3.33\%, and 3.52\% to 1.89\%, 2.53\%, and
2.85\%. Mean decoder time across the four tasks is 173.78 ms for ordinary
sampling and 174.68 ms for Critical-Offset sampling.

\subsubsection{Critical-Offset sensitivity}

We vary two parameters while holding the total budget at $B=100$ rollouts
per POMO start and augmentation. The critical-offset rollout ratio $\rho$
assigns $B_{\mathrm{off}}=\rho B$ rollouts to nonzero offsets and leaves the
rest at $\alpha=0$. The offset-strength spacing $\Delta\alpha$ defines
$\Lambda_{\Delta\alpha}=\{\Delta\alpha,2\Delta\alpha,\ldots,1.0\}$; its
values are assigned cyclically to the offset rollouts. We test
$\rho\in\{0.05,0.10,0.20,0.50\}$ and
$\Delta\alpha\in\{0.05,0.10,0.20,0.50\}$, excluding three combinations with
$B_{\mathrm{off}}<|\Lambda_{\Delta\alpha}|$. As in the decoding ablation,
reference masks isolate decoder behavior, while critical nodes are extracted
from LangNCO-Gemma4 DSL. The setting used in the main paper is
$(\rho,\Delta\alpha)=(0.10,0.20)$.

\begin{table*}[t]
\caption{Critical-Offset sensitivity at a fixed budget of 100 rollouts per
POMO start and augmentation, with rollout ratio $\rho$ by row and strength
spacing $\Delta\alpha$ by column.
Entries are mean reference gap $\pm$ standard deviation over feasible
solutions; every filled cell attained 300/300 feasible solutions. Dashes mark
combinations that cannot cover every strength at least once. Bold denotes the
lowest gap for each dataset, and $\dagger$ marks the setting used in the main
paper.}
\label{tab:critical-offset-sensitivity}
\centering
\footnotesize
\begin{tabular}{@{}lcrrrr@{}}
\toprule
Dataset & $\rho$ & \multicolumn{4}{c}{Reference gap [\%] at offset-strength spacing $\Delta\alpha$} \\
\cmidrule(lr){3-6}
& & 0.05 & 0.10 & 0.20 & 0.50 \\
\midrule
Road & 0.05 & \NA & \NA & $1.63\pm1.66$ & $1.63\pm1.66$ \\
& 0.10 & \NA & $1.63\pm1.66$ & $1.63\pm1.65^{\dagger}$ & $1.64\pm1.67$ \\
& 0.20 & $1.64\pm1.65$ & $\mathbf{1.62}\pm1.62$ & $1.65\pm1.67$ & $1.65\pm1.67$ \\
& 0.50 & $1.66\pm1.62$ & $1.66\pm1.63$ & $1.65\pm1.59$ & $1.68\pm1.66$ \\
\addlinespace
Time & 0.05 & \NA & \NA & $1.94\pm3.02$ & $1.94\pm3.03$ \\
& 0.10 & \NA & $1.89\pm2.97$ & $1.89\pm3.00^{\dagger}$ & $1.88\pm2.97$ \\
& 0.20 & $1.87\pm2.96$ & $1.87\pm2.96$ & $1.86\pm2.96$ & $1.86\pm2.94$ \\
& 0.50 & $1.86\pm2.94$ & $1.86\pm2.95$ & $\mathbf{1.85}\pm2.95$ & $1.86\pm2.95$ \\
\addlinespace
Vehicle & 0.05 & \NA & \NA & $2.60\pm2.69$ & $2.63\pm2.71$ \\
& 0.10 & \NA & $2.47\pm2.63$ & $2.53\pm2.65^{\dagger}$ & $2.52\pm2.69$ \\
& 0.20 & $2.34\pm2.49$ & $2.37\pm2.59$ & $2.41\pm2.64$ & $2.42\pm2.63$ \\
& 0.50 & $\mathbf{2.26}\pm2.47$ & $2.30\pm2.58$ & $2.33\pm2.63$ & $2.38\pm2.66$ \\
\addlinespace
Order & 0.05 & \NA & \NA & $2.89\pm3.35$ & $2.88\pm3.28$ \\
& 0.10 & \NA & $2.74\pm3.15$ & $2.85\pm3.30^{\dagger}$ & $2.86\pm3.28$ \\
& 0.20 & $2.68\pm3.00$ & $2.63\pm2.91$ & $2.71\pm3.06$ & $2.81\pm3.19$ \\
& 0.50 & $\mathbf{2.59}\pm2.84$ & $2.60\pm2.90$ & $2.72\pm3.00$ & $2.74\pm3.01$ \\
\bottomrule
\end{tabular}
\end{table*}

All 13 evaluated settings reach 100\% feasibility on all four datasets. In
particular, $\rho=0.05$ uses only five offset-guided rollouts, yet both
evaluable spacings recover a feasible solution for every instance. The
feasibility gains therefore do not depend on a narrow allocation or strength
grid: mixing a small number of directed trajectories into ordinary sampling
is already sufficient to avoid the observed future deadlocks.

Unlike feasibility, conditional solution quality does vary across the
parameter grid. At $\Delta\alpha=0.20$, increasing $\rho$ from 0.05 to 0.50
lowers the Vehicle gap from 2.60\% to 2.33\% and the Order gap from 2.89\% to
2.72\%; Time improves more modestly from 1.94\% to 1.85\%. This indicates that
allocating more of the fixed budget to guided trajectories can improve the
best feasible route found, even after feasibility has saturated at 100\%.

The effect of $\Delta\alpha$ is task-dependent rather than uniformly smaller.
On Order, changing $\Delta\alpha$ at fixed $\rho$ produces up to a 0.18-point
gap range, comparable to the effect of changing $\rho$ at fixed
$\Delta\alpha$. At $\rho=0.50$, the two finer grids give 2.59--2.60\%,
versus 2.74\% for
$\Delta\alpha=0.50$. Thus, $\rho$ controls how many guided candidates are
available, whereas $\Delta\alpha$ controls the coverage and resolution of
intervention strengths. Both can matter for finding a route that remains
aligned with the base policy while satisfying the constraint.

The task pattern suggests a mechanism for this quality effect. Under ordinary
hard-masked sampling, Vehicle and Order exhibit both more deadlocks and higher
conditional gaps than Road and Time (Table~\ref{tab:decoding-ablation}). Hard
masking guarantees only immediate legality: in Vehicle, other customers can
consume capacity needed by assigned nodes, while in Order, a depot return can
close a route before all order-linked customers are completed. Once such a
future consequence becomes immediate, the mask can abruptly remove the
policy's preferred actions. If legal actions remain, renormalization forces
choices that the learned policy originally considered unlikely, disrupting
its distance-minimizing route structure through a costly detour; otherwise,
the rollout deadlocks. A larger $\rho$ creates more trajectories that
prioritize critical nodes before this point, while a finer $\Delta\alpha$ grid
includes weak offsets that preserve the base policy as well as strong offsets
that avert deadlock. Selecting among these trajectories can therefore recover
a policy-aligned route rather than merely a feasible one. Time also creates a
future dependency, but each instance has a single windowed target rather than
multiple coupled critical nodes, which may explain its smaller variation. In
contrast, Road's forbidden edge is removed immediately, creates no comparable
future commitment, and its gap remains within 1.62--1.68\% across the grid.

The best four-task mean, 2.09\% at $(0.50,0.05)$, is only 0.13 percentage
points below the main setting's 2.23\%. The parameters therefore offer useful
quality control, with task-dependent effects, while the main
result---100\% feasibility with near-best quality while retaining 90 ordinary
rollouts---does not depend on a narrow parameter choice.

\subsection{DSL versus Free-Form Python}

Table~\ref{tab:dsl-python} compares the two representation-specific generation
pipelines; the LLM, sentence, downstream solver, and Critical-Offset Decoding
are fixed. Failure is
the number of instances without a feasible final output, and gap is averaged
only over feasible outputs.

\begin{table*}[t]
\caption{Schema-validated DSL versus free-form Python rule generation. Each
row contains 300 instances.}
\label{tab:dsl-python}
\centering
\small
\begin{tabular}{@{}lllrrr@{}}
\toprule
Dataset & Generator & Representation & Feasible & Failure & Gap [\%] \\
\midrule
Road & Gemma4 & DSL & 300/300 (100.00\%) & 0 & 1.63 \\
Road & Gemma4 & Python & 300/300 (100.00\%) & 0 & 1.63 \\
Road & GPT-5 & DSL & 300/300 (100.00\%) & 0 & 1.63 \\
Road & GPT-5 & Python & 300/300 (100.00\%) & 0 & 1.63 \\
Time & Gemma4 & DSL & 300/300 (100.00\%) & 0 & 1.89 \\
Time & Gemma4 & Python & 289/300 (96.33\%) & 11 & 1.89 \\
Time & GPT-5 & DSL & 300/300 (100.00\%) & 0 & 1.89 \\
Time & GPT-5 & Python & 300/300 (100.00\%) & 0 & 1.92 \\
Vehicle & Gemma4 & DSL & 300/300 (100.00\%) & 0 & 2.53 \\
Vehicle & Gemma4 & Python & 300/300 (100.00\%) & 0 & 2.53 \\
Vehicle & GPT-5 & DSL & 300/300 (100.00\%) & 0 & 2.68 \\
Vehicle & GPT-5 & Python & 299/300 (99.67\%) & 1 & 2.62 \\
Order & Gemma4 & DSL & 300/300 (100.00\%) & 0 & 3.26 \\
Order & Gemma4 & Python & 268/300 (89.33\%) & 32 & 3.05 \\
Order & GPT-5 & DSL & 300/300 (100.00\%) & 0 & 2.97 \\
Order & GPT-5 & Python & 300/300 (100.00\%) & 0 & 3.30 \\
\bottomrule
\end{tabular}
\end{table*}

The DSL is feasible in all eight generator--task conditions. Python falls
below 100\% in Gemma4-Time, Gemma4-Order, and GPT-5-Vehicle, reaching a minimum
of 89.33\%. Its occasionally smaller feasible-output gap reflects selection
after failures and does not establish better solution quality.

\subsection{Pairwise Constraint Results}

Table~\ref{tab:pairwise} compares Gemma4 separate and joint generation. No
pair-specific prompt or handler is used. OR-Tools (GCA) is a 600-second
time-limited reference. Feasibility is counted independently for each method
over all 300 instances; each LangNCO gap is computed only on the intersection
where that row and GCA are both feasible. Therefore, objective columns can
summarize different feasible subsets.

\begin{table*}[t]
\caption{All pairwise constraint results. Gap is relative to the OR-Tools GCA
reference on the jointly feasible subset.}
\label{tab:pairwise}
\centering
\footnotesize
\begin{tabular}{@{}llrrr@{}}
\toprule
Combination & Method & Feasible & Gap vs. GCA [\%] & Objective \\
\midrule
Road + Time & OR-Tools (GCA) & 300/300 (100.00\%) & \NA & $9{,}875.92\pm4{,}773.63$ \\
Road + Time & Separate & 300/300 (100.00\%) & $2.25\pm3.45$ & $10{,}016.40\pm4{,}679.15$ \\
Road + Time & Joint & 297/300 (99.00\%) & $2.27\pm3.46$ & $9{,}996.15\pm4{,}696.13$ \\
\addlinespace
Road + Vehicle & OR-Tools (GCA) & 300/300 (100.00\%) & \NA & $9{,}998.89\pm4{,}844.93$ \\
Road + Vehicle & Separate & 300/300 (100.00\%) & $3.14\pm2.82$ & $10{,}253.60\pm4{,}813.32$ \\
Road + Vehicle & Joint & 300/300 (100.00\%) & $3.14\pm2.82$ & $10{,}253.60\pm4{,}813.32$ \\
\addlinespace
Road + Order & OR-Tools (GCA) & 300/300 (100.00\%) & \NA & $9{,}855.80\pm4{,}812.87$ \\
Road + Order & Separate & 300/300 (100.00\%) & $3.94\pm4.33$ & $10{,}152.61\pm4{,}749.32$ \\
Road + Order & Joint & 274/300 (91.33\%) & $4.48\pm4.80$ & $10{,}060.28\pm4{,}734.18$ \\
\addlinespace
Time + Vehicle & OR-Tools (GCA) & 300/300 (100.00\%) & \NA & $10{,}074.32\pm4{,}841.97$ \\
Time + Vehicle & Separate & 300/300 (100.00\%) & $4.01\pm3.71$ & $10{,}384.94\pm4{,}776.70$ \\
Time + Vehicle & Joint & 300/300 (100.00\%) & $4.01\pm3.71$ & $10{,}384.94\pm4{,}776.70$ \\
\addlinespace
Time + Order & OR-Tools (GCA) & 289/300 (96.33\%) & \NA & $9{,}981.26\pm4{,}764.97$ \\
Time + Order & Separate & 292/300 (97.33\%) & $5.18\pm5.17$ & $10{,}369.51\pm4{,}666.54$ \\
Time + Order & Joint & 286/300 (95.33\%) & $4.82\pm4.96$ & $10{,}388.42\pm4{,}684.52$ \\
\addlinespace
Vehicle + Order & OR-Tools (GCA) & 297/300 (99.00\%) & \NA & $10{,}138.43\pm4{,}853.72$ \\
Vehicle + Order & Separate & 294/300 (98.00\%) & $6.77\pm5.80$ & $10{,}603.09\pm4{,}671.41$ \\
Vehicle + Order & Joint & 296/300 (98.67\%) & $6.62\pm5.59$ & $10{,}676.40\pm4{,}788.87$ \\
\bottomrule
\end{tabular}
\end{table*}

Separate, Joint, and GCA return respectively 1786, 1753, and 1786 feasible
solutions out of 1800. Joint generation is therefore not uniformly superior.
It recovers two Vehicle + Order instances by integrating a shared node's
vehicle assignment with same-vehicle ordering, but it fails on 26 Road + Order
instances, illustrating that combined context can also lose individual
semantics.

\begin{table*}[t]
\caption{OR-Tools pairwise feasibility by first-solution strategy.
PCI: parallel cheapest insertion; LCI: local cheapest insertion; PMCA: path
most constrained arc; GCA: global cheapest arc; SV: savings; AT: automatic.}
\label{tab:ortools-strategy}
\centering
\small
\begin{tabular}{@{}lrrrrrr@{}}
\toprule
Combination & \multicolumn{6}{c}{Feasibility [\%]} \\
\cmidrule(l){2-7}
& PCI & LCI & PMCA & GCA & SV & AT \\
\midrule
Road + Time & 100.00 & 100.00 & 100.00 & 100.00 & 100.00 & 100.00 \\
Road + Vehicle & 100.00 & 100.00 & 100.00 & 100.00 & 24.67 & 73.67 \\
Road + Order & 98.67 & 98.67 & 8.33 & 100.00 & 9.33 & 11.67 \\
Time + Vehicle & 99.00 & 96.33 & 99.67 & 100.00 & 22.33 & 73.33 \\
Time + Order & 90.00 & 86.67 & 7.33 & 96.33 & 6.67 & 13.00 \\
Vehicle + Order & 96.33 & 88.00 & 14.00 & 99.00 & 4.00 & 17.67 \\
\midrule
Total feasible & 1752/1800 & 1709/1800 & 988/1800 & \textbf{1786/1800} & 501/1800 & 868/1800 \\
\bottomrule
\end{tabular}
\end{table*}

GCA provides the greatest aggregate reference coverage and is therefore used
for pairwise gaps. This selection increases time-limited reference coverage;
it is not a claim of general superiority for OR-Tools or LangNCO.

\subsection{Additional Constraints}

\begin{table*}[t]
\caption{Feasibility on twelve additional constraint types, 50 instances
each. No constrained objective reference is available for these datasets.}
\label{tab:additional}
\centering
\small
\begin{tabular}{@{}lrr@{}}
\toprule
Constraint type & NCO w/o intervention & LangNCO-Gemma4 \\
\midrule
First delivery & 0/50 (0.00\%) & 50/50 (100.00\%) \\
Visit position & 4/50 (8.00\%) & 50/50 (100.00\%) \\
Arrival capacity reserve & 33/50 (66.00\%) & 50/50 (100.00\%) \\
Heavy-delivery limit & 5/50 (10.00\%) & 50/50 (100.00\%) \\
Minimum service interval & 1/50 (2.00\%) & 50/50 (100.00\%) \\
Depot closure & 17/50 (34.00\%) & 50/50 (100.00\%) \\
Small-delivery assignment & 0/50 (0.00\%) & 48/50 (96.00\%) \\
Return deadline & 6/50 (12.00\%) & 44/50 (88.00\%) \\
Vehicle workload limit & 8/50 (16.00\%) & 50/50 (100.00\%) \\
Consecutive deliveries & 1/50 (2.00\%) & 50/50 (100.00\%) \\
Customer incompatibility & 13/50 (26.00\%) & 50/50 (100.00\%) \\
Restricted-area transition & 6/50 (12.00\%) & 50/50 (100.00\%) \\
\midrule
\textbf{Total} & \textbf{94/600 (15.67\%)} & \textbf{592/600 (98.67\%)} \\
\bottomrule
\end{tabular}
\end{table*}

The frozen policy without intervention is feasible on 15.67\% overall.
LangNCO-Gemma4 reaches 98.67\% and is perfect on ten tasks. The two
small-delivery failures concern critical-node extraction from
attribute-defined targets, whereas the six return-deadline failures concern a
semantic rule-generation error. Section~\ref{sec:failures} separates these
mechanisms.

\section{Failure and Mechanistic Analyses}
\label{sec:failures}

\subsection{Free-Form Python Failures}

Free-form failures concentrate in low-level tensor integration rather than in
recognizing the requested condition. Gemma4 produces 11 Time and 32 Order
runtime failures; GPT-5 produces one Vehicle implementation failure. Three
representative mechanisms follow.

\paragraph{Shape mismatch.}
A Time rule adds \texttt{curr\_time} of shape $[B]$ directly to
\texttt{travel\_times} of shape $[B,N]$, causing a broadcasting error. The
rule needs only the designated destination and should jointly index the batch,
current node, and target to obtain a $[B]$ pairwise value.

\paragraph{Device mismatch.}
An Order rule initializes
\begin{lstlisting}[language=Python]
met = torch.ones(can_visit.shape[0], dtype=torch.bool)
\end{lstlisting}
on the CPU, while \texttt{visited\_by}, \texttt{vehicle\_idx}, and
\texttt{can\_visit} reside on the GPU. A subsequent Boolean operation fails
unless the new tensor is created on \texttt{can\_visit.device}.

\paragraph{Copy-only advanced indexing.}
A Vehicle rule uses
\begin{lstlisting}[language=Python]
can_visit[rows][:, cols] = False
\end{lstlisting}
The first Boolean index creates a temporary copy, so the assignment does not
modify \texttt{can\_visit}. This silent failure requires a single direct index
into the original tensor. The DSL avoids all three classes by leaving shapes,
devices, broadcasting, and in-place mask composition to the compiler.

\subsection{Pairwise and Additional-Constraint Failures}

Separate Vehicle + Order generation can produce individually plausible rules
whose conjunction loses a cross-constraint implication. For example:
\begin{lstlisting}[style=constraint]
- Nodes 30, 3 must be served by Vehicle ID 5.
- Nodes 30, 60 must be served before Node 57 by the same vehicle.
\end{lstlisting}
The Vehicle rule restricts only nodes 30 and 3. The Order rule blocks 57 until
30 and 60 have been visited by the current vehicle, but independent generation
does not infer that the shared node 30 makes vehicle 5 the required vehicle for
60 and 57 as well. If another vehicle serves 60 first, vehicle 5 cannot meet
the route-local prerequisite. Separate generation fails on two such instances;
Joint constrains all four related nodes to vehicle 5 and completes both.

Joint generation is not uniformly better. One Road + Time rule checks whether
\texttt{current\_time} itself lies in the window rather than adding travel time
to the target. Several Road + Order and Time + Order rules omit
\texttt{by: vehicle\_idx} from \texttt{all\_visited}; this represents global
precedence but loses the requested same-vehicle condition. The frequency of
the latter error in Road + Order may indicate interference from the road's
node-to-node relation, but that interpretation is an inference from outputs,
not a causal finding. Joint Vehicle + Order also sometimes misses the same
cross-constraint implication. Moreover, an OR-Tools no-solution result under
the time limit is not proof of infeasibility, so we assign mechanistic causes
only when the generated rule or trace directly identifies one.

Small-delivery assignment exposes a different limitation. The Rule Executor
correctly evaluates a \texttt{where} target such as ``demand at most 4'' at
every step, but the static critical-node extractor resolves only literal IDs.
The designated route can return before serving every member of the
attribute-defined set, after which other vehicles are blocked from them.

For diagnosis, we preserve the same hard constraint and replace one failed
instance's attribute predicate
\begin{lstlisting}[style=constraint]
Vehicle ID 0 must serve all and only customer nodes whose demand is at most 4.
\end{lstlisting}
with its instance-grounded equivalent:
\begin{lstlisting}[style=constraint]
Vehicle ID 0 must serve all and only customer nodes
5, 8, 10, 12, 18, 19, 27, 29, 41, 42, 46, 47, 57, 58, 60, 62, 63, 67, 68.
\end{lstlisting}
The original form yields no solution, whereas the literal form extracts the
critical set and completes under the same hard semantics. This controlled
reformulation directly supports extractor coverage as the cause. Static
attributes could be expanded deterministically once per instance without
changing the Rule Executor, offset equation, or frozen policy; aggregate
results with that extension remain future work.

Return deadline instead exposes semantic generation error. For ``Vehicle 1
must arrive at the depot by time 25,'' a failed rule blocks late arrivals to
\texttt{all\_customers} but never targets the depot-return action. The rule
discourages new service after the deadline without enforcing the required
return. Residual errors therefore separate into missing cross-constraint
implications, incorrect state or same-vehicle semantics, and static
critical-node extraction coverage.

\subsection{Critical-Offset Sampling Case Study}

We compare an Ordinary Sampling rollout that deadlocks with a Critical-Offset
Sampling rollout that completes on the same instance and sample index. Raw
probability refers to the frozen policy probability after hard masking, while
the offset weight is $q_t^{(\alpha)}$ from Equation~\ref{eq:offset-supp}.

\subsubsection{Time: preserving a service opportunity}

Both rollouts first move from the depot to node 14. Under Ordinary Sampling,
node 2 remains legal but has raw probability $1.34\times10^{-9}$; selecting
node 13 makes its time window unreachable and the rollout deadlocks. In the
Critical-Offset rollout, node 2 has raw probability $5.69\times10^{-9}$ but
offset weight 1.00, so it is served immediately and the rollout completes.

The hard mask cannot prohibit node 13 because it is locally legal. The offset
adds a trajectory that uses the expiring service opportunity.

In this representative case, a critical node is legal but receives negligible policy
probability. The Rule Executor has no basis to remove a different action that
is legal at the current step. Critical-Offset Sampling reallocates part of the
existing rollout budget toward constraint-referenced nodes without relaxing
masks. The lower aggregate gaps for Critical-Offset Sampling in
Table~\ref{tab:decoding-ablation} are consistent with this interpretation, but
the case study does not establish it causally. The method remains a search bias,
not a look-ahead proof of completion, and its current fixed-ID coverage is the
limitation diagnosed above.